%% file: main.tex
\documentclass[conference]{IEEEtran}
\usepackage{times}

\usepackage[numbers]{natbib}
\usepackage{multicol}
\usepackage{xcolor}
\usepackage[bookmarks=true]{hyperref}

\usepackage{enumitem}

\newcommand{\tup}[1]
           {
             \relax\ifmmode
             \langle #1 \rangle
             \else $\langle$ #1 $\rangle$ \fi
           }
\usepackage{amsmath}
\newcommand{\num}[1]{\relax\ifmmode \mathbb #1\else $\mathbb #1$\fi}
\newcommand{\nnnum}[1]{\relax\ifmmode 
  {\mathbb #1}_{\geq 0} \else ${\mathbb #1}_{\geq 0}$
  \fi}

\usepackage{amssymb}
\input{sections/0-preamble.tex}

\begin{document}

\title{Safety and progress proofs for a reactive planner and controller for autonomous driving}



\author{\IEEEauthorblockN{Abolfazl Karimi}
\IEEEauthorblockA{\textit{Department of Computer Science} \\
\textit{University of North Carolina}\\
Chapel Hill, United States \\
ak@cs.unc.edu}
\and
\IEEEauthorblockN{Manish Goyal}
\IEEEauthorblockA{\textit{Department of Computer Science} \\
\textit{University of North Carolina}\\
Chapel Hill, United States \\
manishg@cs.unc.edu}
\and
\IEEEauthorblockN{Parasara Sridhar Duggirala}
\IEEEauthorblockA{\textit{Department of Computer Science} \\
\textit{University of North Carolina}\\
Chapel Hill, United States \\
psd@cs.unc.edu}
}


%

\maketitle

\import{sections/}{0-abstract.tex}

\IEEEpeerreviewmaketitle

\import{sections/}{1-introduction.tex}
\import{sections/}{2-related_work.tex}
\import{sections/}{4-0-preliminaries.tex}

\import{sections/}{5-0-percplan-cntrl.tex}

\import{sections/}{6-0-rigorous-analysis.tex}

\import{sections/}{8-0-evaluation.tex}

\import{sections/}{9-conclusion.tex}


\bibliographystyle{plainnat}
\bibliography{main}
\appendix
\import{sections/}{appendix}

\end{document}

%% file: sections/0-preamble.tex
\usepackage{import}
\usepackage{graphicx}
\usepackage{algorithm}
\usepackage{algpseudocode}
\usepackage{subfigure}
\usepackage{hyperref}

\usepackage{amsthm}
\newtheorem{theorem}{Theorem}
\newtheorem{lemma}[theorem]{Lemma}
\newtheorem{definition}[theorem]{Definition}

%% file: sections/0-abstract.tex
\begin{abstract}
In this paper, we perform safety and performance analysis of an autonomous vehicle that implements \emph{reactive planner and controller} for navigating a race lap.
Unlike traditional planning algorithms that have access to a map of the environment, reactive planner generates the plan purely based on the current input from sensors.
Our reactive planner selects a waypoint on the local Voronoi diagram and we use a pure-pursuit controller to navigate towards the waypoint.
Our safety and performance analysis has two parts.
The first part demonstrates that the reactive planner computes a plan that is locally consistent with the Voronoi plan computed with full map.
The second part involves modeling of the evolution of vehicle navigating along the Voronoi diagram as a hybrid automata.
For proving the safety and performance specification, we compute the reachable set of this hybrid automata and employ some enhancements that make this computation easier.
We demonstrate that an autonomous vehicle implementing our reactive planner and controller is safe and successfully completes a lap for five different circuits.
In addition, we have implemented our planner and controller in a simulation environment as well as a scaled down autonomous vehicle and demonstrate that our planner works well for a wide variety of circuits.

\end{abstract}

%% file: sections/1-introduction.tex
\section{Introduction}

A typical workflow of an autonomous vehicle involves three main steps: perception, planning, and control.
%
%
%
%
The estimation of the current state of the vehicle and its surroundings is performed by perception.
The sequence of states to be visited for safely navigating the environment is determined by planning.
And finally, the plan is realized through a control algorithm.
Crucially, these three steps assume that the vehicle has access to an accurate map of the environment that is either obtained from simultaneous localization and mapping or provided by the user.

In contrast to the above mentioned workflow, in this paper, we pursue \emph{reactive planning and control}.
Here, we assume that the map of the environment and the relative position of the vehicle are not known.
Instead, the vehicle generates the plan and control input purely based on the current data that it receives from the sensors.
Hence, the plan and the control inputs are generated \emph{on-the-fly}.
%
We believe that investigating reactive planning and control based approaches are useful in instances where the environment is highly dynamic or mapping of the environment has not yet been performed or is computationally very expensive.
Additionally, the same planning and control mechanisms can be deployed in a wide range of scenarios as it doesn't require any map.
%

One of the main drawbacks of reactive planning and control is that it is very challenging to prove that the safety and progress specifications are satisfied.
Since the map of the environment is not known, the waypoints generated during the motion planning are always relative to the vehicle.
Hence, as the vehicle navigates through the environment, the waypoint in the next instance also evolves.
\emph{This dynamic nature of the state of the vehicle and the waypoints makes the safety analysis very challenging.}
Furthermore, as the vehicle navigates through the environment, the new sensor readings could cause changes in the planned path making safety analysis a very challenging task.

In this paper, we perform safety and performance analysis of a reactive planning and control algorithm deployed on an autonomous vehicle that is navigating a race lap.
Our planner involves computing a Voronoi diagram of the walls visible to the perception and our control algorithm implements the pure-pursuit algorithm.
Our safety and performance analysis has two parts.
In the first part, we demonstrate that the waypoint computed by the reactive planner is consistent with the planner that has access to the full map.
In the second part, we model the co-evolution of the state of the vehicle and the waypoint as a hybrid automata and compute an artifact called \emph{reachable set}.
The reachable set contains all the configurations visited by the autonomous vehicle while realizing the plan using the pure-pursuit control algorithm.
We show that the reachable set is safe (no overlap with the boundaries of the race track) and achieves a fixed point after the vehicle completes a full lap.
This proves that the vehicle satisfies the safety specification while guaranteeing that it will eventually complete the lap.
%
%
We also employ abstraction techniques from hybrid systems literature~\cite{tiwari2002series,prabhakar2015hybrid} to improve the efficiency of the reachable set computation algorithm.
%

%


The primary contribution of this paper is to establish the safety and performance specification of reactive planning and control algorithm used for navigation of autonomous vehicles.
Unlike many approaches that strictly investigate safety of planning or safety of closed loop control behavior, we consider both these aspects at the same time for proving safety.
%
%
%
In addition, we also show the effectiveness of our reactive planning and control algorithm in two ways.
First, we develop a simulation environment of a race lap using Unreal engine and deploy the vehicle in various types of race tracks.
Second, we implement the algorithm on an open source hardware platform of F1Tenth, a scale down version of autonomous vehicle built on Traxxas RC car.
%


%% file: sections/2-related_work.tex
\section{Related work}
\label{sec:relwork}



Planning and control of autonomous vehicles as well as their safety verification has received a lot of attention in the recent past. A comprehensive survey of many of these techniques are provided in~\cite{paden2016survey,machines5010006}. In this section, we briefly discuss the various planning and control techniques and some of the verification techniques presented in the literature and contrast it with the techniques presented in this paper.


Planners for autonomous vehicles are often hierarchical in nature~\cite{miller2008team,urmson2008autonomous}. The task planner selects the high level task to be performed by the vehicle and the motion planner implements the task decided by the task planner. In this paper, since we are concerned with an autonomous racing vehicle, the task of the vehicle is to complete the lap. We hence focus our attention to the motion planning aspect of the vehicle. 

Motion planning of autonomous vehicles is primarily divided into two methods. First are the geometric based planning methods where the sequence of waypoints for the vehicle are decided on the geometry of the configuration space~\cite{thrun2006stanley,montemerlo2008junior}. Planning based on Voronoi diagrams is one of the popular geometric techniques for planning~\cite{takahashi1989motion,shkolnik2009reachability}.
Second are the sampling based motion planning techniques.
In particular, RRTs~\cite{lavalle1998rapidly,karaman2011sampling} and PRMs~\cite{kavraki1996probabilistic} are two of the most influential techniques. In the literature, several variants of RRTs and PRMs specific to the domain of autonomous vehicles have been proposed~\cite{xu2012real,aoude2010threat,levinson2011towards,perez2012lqr,kant1986toward}. All of these techniques assume that a partial map of the environment is provided and the location of the vehicle in the map is known. In the case of reactive planning and control, we do not assume that a partial map of the system is known.

Reactive planning was proposed as an alternative to offline planning when all the information for completing a task are not available to the robot~\cite{georgeff1987reactive}, or when the environment is highly dynamic~\cite{belkhouche2009reactive} such as robot soccer~\cite{bruce2002real} or human collaborative environment~\cite{dumonteil2015reactive}.
Such reactive plans have been successfully deployed in robots that are resource constrained such as small-scale helicopters~\cite{redding2007real}, micro air vehicles~\cite{sharma2012reactive}, and autonomous sailboats~\cite{petres2011reactive}.
Additionally, reactive planners have also been used to modify an existing plan due to the presence of dynamic obstacles~\cite{moreau2019reactive,moreau2019reactive-be}.
Finally, when autonomous vehicles have to satisfy service requests along liveness specification given according to a temporal logic formula, a reactive sampling based motion planning algorithm has been used~\cite{vasile2014reactive,vasile2020reactive}. 

Control of autonomous vehicles are also divided into two categories, geometric and model based. In geometric control techniques, the underlying geometric properties of the bicycle model are used in order to make the vehicle reach its destination. Two popular geometric control techniques are pure-pursuit~\cite{park2014development,Snider.2009} and Stanley~\cite{hoffmann2007autonomous}. These techniques are intuitive and easy to implement. 
Model based control techniques assume a given model of the vehicle and generate control inputs depending on the model~\cite{de1998feedback,murray1993nonholonomic}. Model predictive control based methods that are path tracking, unconstrained, and with dynamic car model have been proposed in the literature~\cite{falcone2007predictive,falcone2008linear,raffo2009predictive}. In fact, MPC based control methods have been used in other autonomous racing vehicles~\cite{talvala2011pushing}.

In safety analysis of mobile robots, synthesis of safe plan based on temporal logic specification has received a lot of attention~\cite{kress2009temporal,fainekos2009temporal,Kloetzertemp}. For proving the safety of the control algorithms, various reachable set computation methods have been proposed and evaluated for behavior of an autonomous vehicle at different scenarios~\cite{AlthoffDolan,LygerosSastry,ACCVerified,ACCTwoApproaches}. 
%
%
%
Given a map of the environment, accurate sensors, noise free localization, and accurate model of the vehicle dynamics, it is possible to provide a sequence of waypoints for moving along the track and prove that the vehicle finishes the lap without colliding with any obstacles using standard reachability based techniques. 
However, in this paper, we consider reactive planning and control, where the waypoint dynamically changes along with the vehicle position and orientation.


The works that are closest to the current work are~\cite{Das.2011} and ~\cite{Dolgov.2010}. In both these works, the plan of the vehicle is based on computing the Voronoi diagram. In~\cite{Das.2011}, the authors do not precisely model the interaction between the waypoint decided by the planner and the control algorithm and hence do not provide any safety guarantees even when the track is known apriori. In~\cite{Dolgov.2010}, in contrast to our work, the authors assume that a partial map of the environment is known. Furthermore, they do not provide any safety guarantees assuming an uncertainty in the initial conditions of the vehicle.

%% file: sections/4-0-preliminaries.tex
\section{Preliminaries}

\subimport{}{4-4-problemdefn.tex}
\subimport{}{4-2-purepursuit.tex}

%% file: sections/4-4-problemdefn.tex
\subsection{Problem definition}
Consider that an autonomous vehicle is tasked to complete a circuit track without hitting the borders of the track.
Completing the track is a \emph{progress} requirement, and avoiding the track boundaries is a \emph{safety} requirement.
Furthermore, the planning and control algorithms are required to be \emph{reactive}: at each time step, steering and speed controls are calculated from the current sensor data only.
Assuming that there is no lateral wheel slippage,
we can isolate the problem of finding the steering control from finding the velocity control.
This is because if a wheel does not slip laterally, it will move along its direction and so the shape (i.e. footprint) of the trajectory only depends on the steering angle.
In other words, the shape of the trajectory is not influenced by the speed, as long as the speed is nonzero.
In this paper, we present a simple reactive planner and controller and formally analyze its safety and progress properties.


%% file: sections/4-2-purepursuit.tex
\subsection{Bicycle model of a car}


\begin{figure}
\centering
\includegraphics[width=45mm]{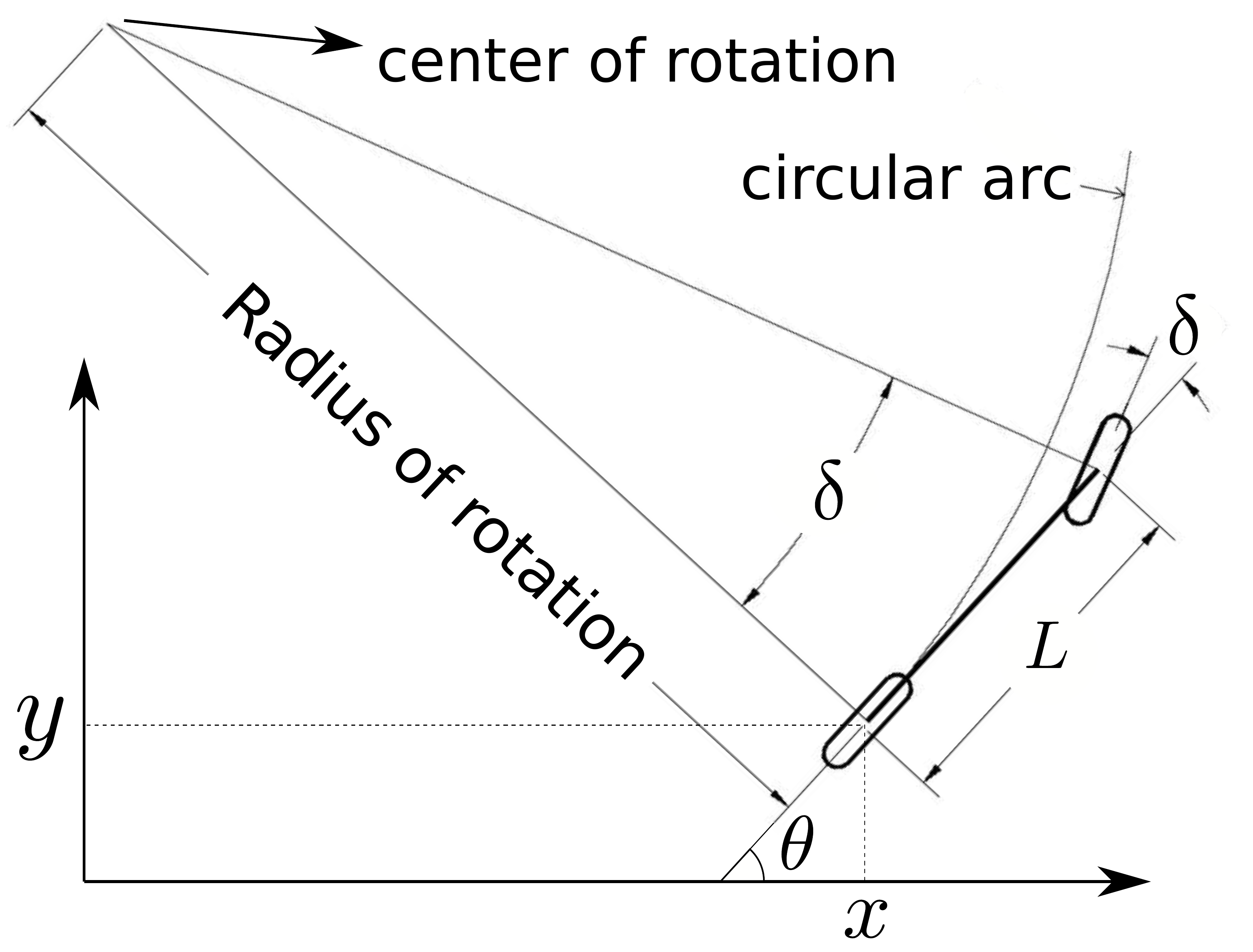}%
\caption{Bicycle model \cite{Snider.2009}}
\label{fig:bicycle}%
\end{figure}

Our formal analysis is based on the
bicycle model of a car, where we imagine that there is one rear wheel at the center of the rear axle and one front wheel at the center of the front axle.
We assume no wheel slippage, and only the front wheel can steer.
This model defines a dynamical system.
The \emph{state} of the system at time $t$
is described by the triple $(x(t), y(t), \theta(t))$
where $(x, y)$ are the coordinates of the rear wheel
in some inertial frame, say the racing track, and
$\theta$ is the angle of the heading direction of the bicycle measured from the $x$-axis counter-clockwise.
If $v$ is the speed (magnitude of velocity) of the rear wheel,
$L$ is the \emph{wheelbase} (distance between the rear and front wheels),
and $\delta$ is the steering angle,
then
\[
\left\{
\begin{array}{l}
     \dot{x} = v \cdot cos(\theta) \\
     \dot{y} = v \cdot sin(\theta) \\
     \dot{\theta} = \frac{\displaystyle v}{L} tan(\delta)
\end{array}
\right.
\]

Illustration of the bicycle model of the vehicle is provided in Figure~\ref{fig:bicycle}.

%% file: sections/5-0-percplan-cntrl.tex
\section{Planning and Control}
In this section we describe our planning and control algorithms which are a variation of  Voronoi-based planning \cite{Das.2011} and geometric control \cite{Snider.2009}.
The safety and progress properties of our algorithms will be formally analyzed and experimentally validated in subsequent sections. 
\label{sec:percp-plan-control}
\input{sections/5-2-planner}
\input{sections/5-3-controller}

\input{sections/5-4-evaluation}

%% file: sections/5-2-planner.tex
\subsection{The Planner}
\label{sec:voronoiplanner}

Our planner is \emph{reactive} in the sense that it does not remember its past inputs or outputs.
The input is simply a 2D point-cloud from lidar, and the output is a 2D \emph{waypoint} passed to the controller.
We assume that the environment is polygonal.
The planner calculates a Voronoi diagram corresponding to the point-cloud, then it chooses a point on the Voronoi diagram as the waypoint.

The first step is to calculate a Voronoi diagram.
Since the environment is polygonal, the planner converts the 2D point cloud to a set of line segments based on co-linearity and connectivity thresholds.
\emph{Co-linearity threshold} determines the minimum angle between consecutive line segments of a polyline.
\emph{Connectivity threshold} determines the minimum distance between two co-linear line segments.
Representing a set of points by a line segment simplifies the computation and representation of the Voronoi diagram.
Since the input to the Voronoi computation is a set of line segments, the Voronoi edges are either linear or parabolic arcs.
After computing the Voronoi diagram, we approximate each parabolic edge by a polyline using a \emph{deviation threshold}.
The \emph{deviation threshold} determines the maximum distance of the points on a parabolic arc from the approximating polyline.
This linear approximation simplifies the planner and its formal analysis.
A Voronoi diagram calculated by the planner is called a \emph{local} Voronoi diagram since it is computed for the point-cloud visible from lidar.
This is in contrast to the \emph{global} Voronoi diagram where the diagram is computed with respect to the whole polygonal environment (which is not available to the car).

The next step is to choose a waypoint based on the local Voronoi diagram.
We choose a point \emph{on} the Voronoi diagram to try to stay as far as possible from the track walls (i.e. to be as safe as possible).
If the waypoint is too close or too far from the vehicle, the controller may make the car steer too sharply or slowly.
We choose among the points at a fixed distance from the center of the rear axle.
This distance is called the \emph{lookahead distance} and the corresponding circle is called the \emph{lookahead circle}.
The lookahead circle may intersect the Voronoi diagram in more than one point, so we need to choose among them.
Since the goal is to make the car progress towards finishing the track, the intersection point further along the heading direction of the car is selected as the waypoint.

%% file: sections/5-3-controller.tex
\subsection{The Controller}
\label{sec:ppcontroller}
The pure-pursuit controller~\cite{Amidi.1991}
as formulated in~\cite{Snider.2009} is used to determine the steering angle.
This simple controller was ``the most stable and accurate tracker"
of the three methods tested in~\cite{Amidi.1991}.
Furthermore, pure-pursuit performed
``fairly well and is quite robust to large errors and discontinuous paths" in comparison to a few more complicated (geometric, kinematic or dynamic) controllers~\cite{Snider.2009}.


\begin{figure}
\centering
\includegraphics[width=60mm]{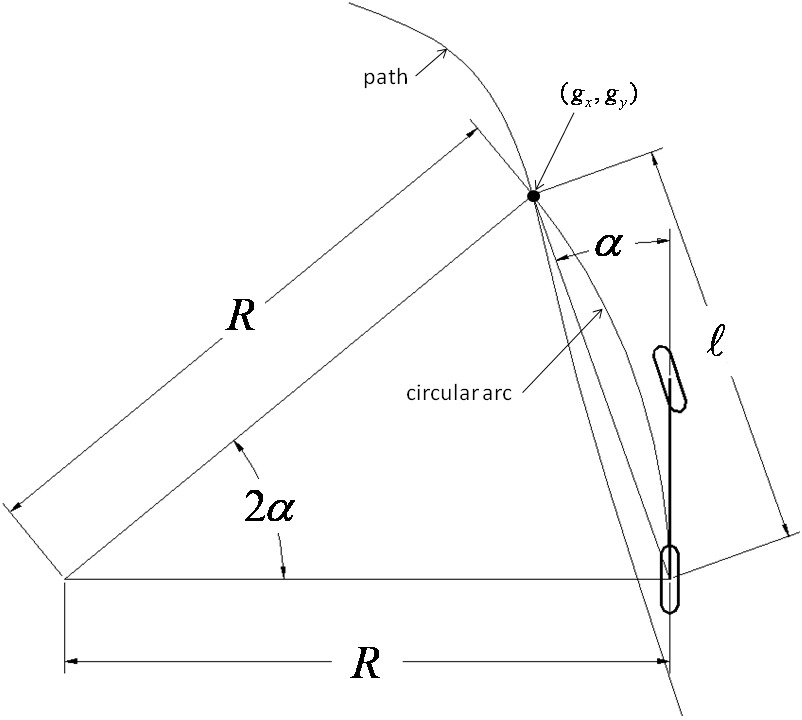}%
\caption{Purepursuit controller \cite{Snider.2009}.}
\label{fig:purepursuit}%
\end{figure}

The input to the pure-pursuit controller is a waypoint described in the car's \emph{rear-axle coordinates}.
The origin of the rear axle coordinates is the center of the rear axle, the x-axis is the heading of the car, and the heading of the y-axis is 90 degrees counterclockwise from the x-axis.
The output is the steering angle $\delta$ for the bicycle model of a car.
Let $(g_x, g_y)$ be the coordinates of the waypoint in the rear-axle frame.
Note that the lookahead distance $\ell$ is $\sqrt{g_x^2+g_y^2}$.
If $L$ is the \emph{wheelbase} of the car (i.e. distance between rear and front axles), then the pure-pursuit steering angle $\delta$ is\footnote{See \cite{Snider.2009} for the derivation of the formula.}
\begin{align}
\delta & = tan^{-1}(\frac{2Lsin(\alpha)}{\ell}) \nonumber \\
 & =  tan^{-1}(\frac{2L g_y}{\ell^2})
 \label{eqn:purepursuit}
\end{align}
Note that this formula is valid even when the waypoint is on the right of the car,
where $\alpha$, $g_y$ and $\delta$ are all negative.
This formula is valid only when $g_x > 0$
i.e. when the waypoint is on the front of the rear axle.
In this case,
we have $\frac{-\pi}{2} < \delta < \frac{\pi}{2} $.
In practice,
$-\delta_{max} \leq \delta \leq \delta_{max} $
where $\delta_{max} < \frac{\pi}{2}$ is the maximum possible angle that the car can steer.
For example,
the maximum steering angle for the Traxxas car in the open source hardware platform of F1Tenth vehicle is about $34$ degrees.



%% file: sections/5-4-evaluation.tex
\section{Evaluation of Planning and Control in Simulation and Noisy Environments}


\subimport{}{5-6-simulation.tex}

\subimport{}{5-7-physical_validation.tex}

%% file: sections/5-6-simulation.tex
\subsection{Computer simulations}

\begin{figure}[!t]
\centering
\includegraphics[width=0.8\columnwidth]{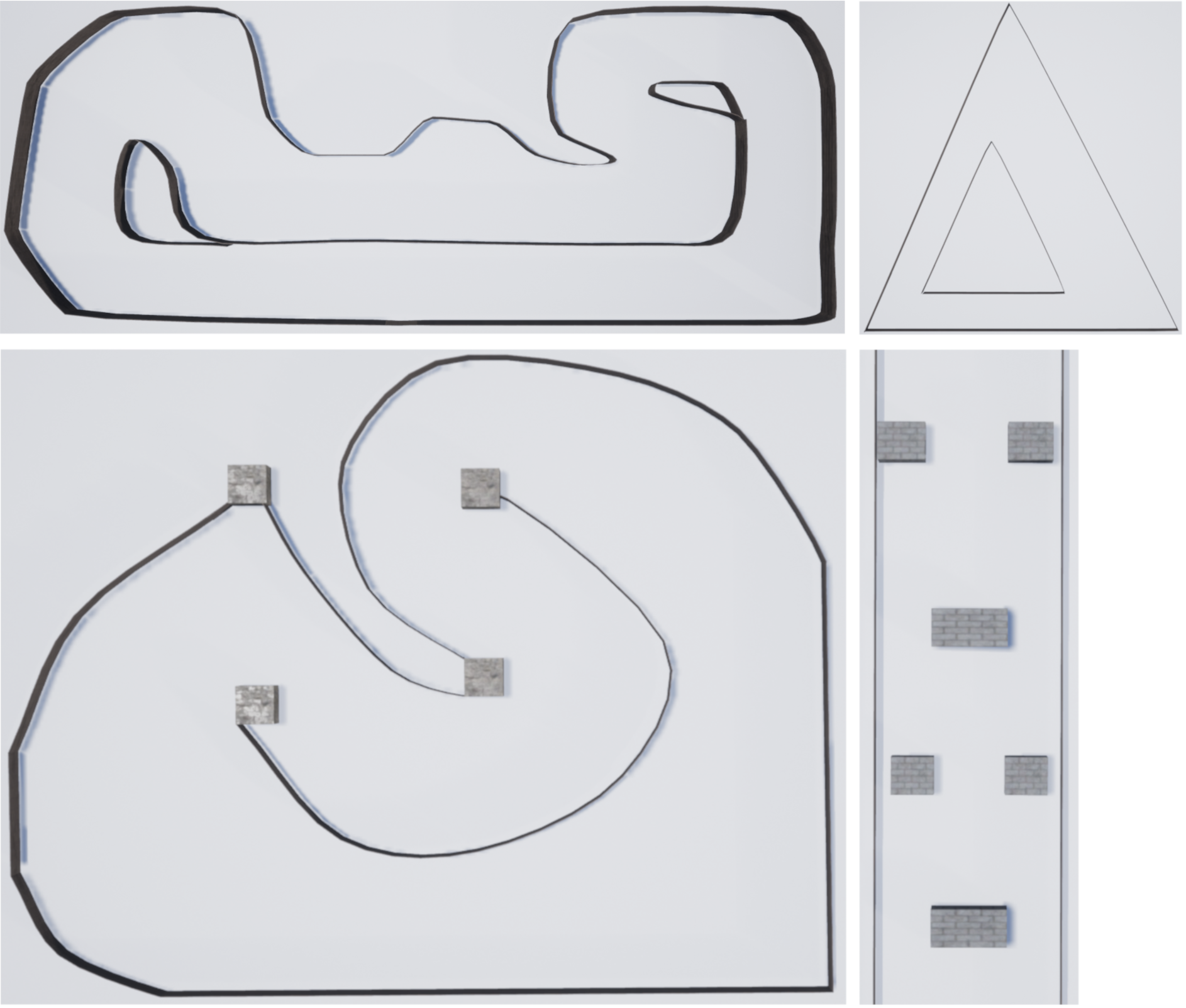}
\caption{Racing tracks for the simulations.}
\label{fig:simulation_tracks}
\end{figure}

We have built a virtual environment including various racing circuits in Unreal Engine for testing the reactive planning and control algorithm.
%
%
Some of these tracks are provided in Figure~\ref{fig:simulation_tracks}.
The tracks in the left column are similar to the tracks on which F1Tenth competitions \cite{f1tenth} were conducted.
Other tracks include a triangular track, and a straight track with static obstacles.
We have also tested the simulation on some polygonal tracks for which we will provide formal analysis in the subsequent sections.
Videos of these simulations are available on the web.\footnote{\url{https://tinyurl.com/ry5xhza}}
In total, of the nine different laps that we have tested, our algorithm was able to successfully able to complete a lap while not colliding with any of the race track boundaries or obstacles.

The vehicle used in the simulations is the vehicle from Unreal Engine 4.21's `Vehicle Advanced' template project.
The default tire friction constants are increased to avoid wheel slippage.
We have implemented a model of Hokuyo UST-10LX LiDAR in Unreal Engine.
The measurements are taken using Unreal Engine's line tracing.
We do not model measurement errors and uncertainty.

%% file: sections/5-7-physical_validation.tex
\subsection{Physical validation}
We also have tested the planner and controller on the open source hardware platform of F1Tenth vehicle.
This is to show the practicality of our simple reactive planning and control in the presence of noise, unstructured environment, and limited computational resource.
For example, Hokuyo UST-10LX LiDAR provides a new point-cloud every 25 milliseconds, but our planning and control is an order of magnitude faster on the Nvidia Jetson TX2's CPU so we can process every point-cloud.

\begin{figure}[!t]
\centering
\includegraphics[width=85mm]{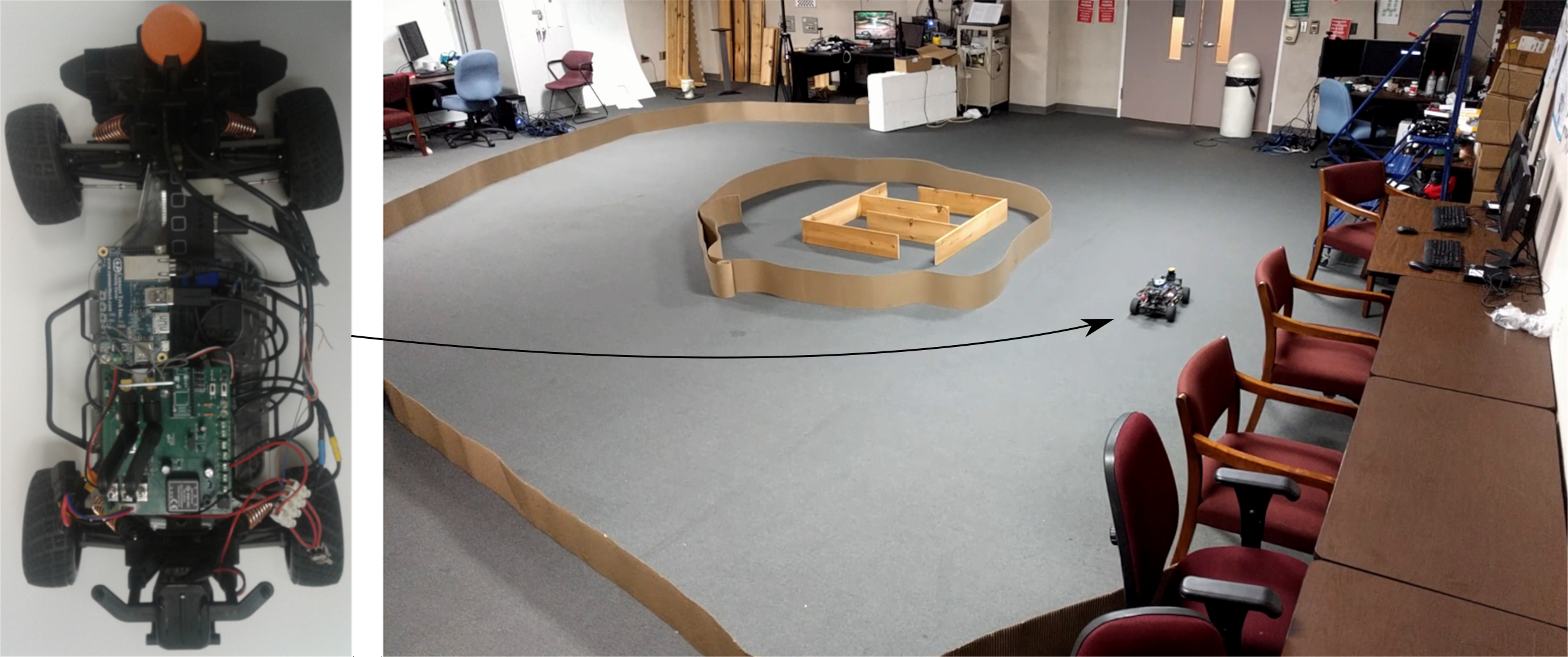}
\caption{Physical validations.}
\label{fig:physical_track}
\end{figure}

We tested the car in two tracks.
The simpler one is shown in Fig.~\ref{fig:physical_track} where the track boundaries is mostly structured using corrugated cardboard rolls.
The harder track (provided in the videos) is an office with chairs and tables around the office walls and in the middle of the office.
Our algorithm manages to successfully complete a lap while avoiding obstacles on both these tracks.

%% file: sections/6-0-rigorous-analysis.tex
\section{Rigorous Analysis of Planning and Control Algorithms for Safety and Progress Properties}
\label{sec:rigorous}

In this section we perform rigorous analysis of the Voronoi planning along with pure-pursuit controller.
We prove that the vehicle would not hit any of the walls and will successfully complete a lap when its starting position belongs to a defined set of initial states.
Our analysis has two parts: the first part is about the properties of the planner and the second part is about the closed loop behavior of the vehicle.

In the first part, we prove that the \emph{local Voronoi diagram} computed by the vehicle from its current environment will be consistent with the \emph{global Voronoi diagram} under certain conditions.
This consistency ensures that the reactive path planning performed by the robot is equivalent to the path planning performed while having access to the map.
The second part of analysis requires computing an artifact called \emph{reachable set} of the closed loop vehicle dynamics.
Given a set of initial positions for the vehicle, the reachable set includes all the set of states visited by the trajectories starting from the set.
We demonstrate that the reachable set does not overlap with any of the walls and achieves a fixed point after finishing the lap, thus proving the safety and progress properties.

\subimport{}{6-1-voronoi.tex}
\import{}{6-2-system_model.tex}
\import{}{6-3-reachable-set.tex}

%% file: sections/6-1-voronoi.tex
\subsection{Consistency of local Voronoi diagrams}
\label{sec:localVoronoi}

In this section, we prove that the Voronoi diagram computed by the vehicle with a given lidar scan is consistent with the global Voronoi diagram computed from the map.
%
%
%
Intuitively, this proof formalizes the notion that a lidar with a sufficiently long range can detect all the edges for computing the Voronoi diagram for a small neighborhood.
%
The proof will formalize the requirements on the track and the range of the lidar.
At each point in time, only a subset of the walls are \emph{visible} to the lidar, i.e. all the walls in its range that are not occluded.
A \emph{local Voronoi diagram} is the Voronoi diagram of the visible walls.
The \emph{global Voronoi diagram} is the Voronoi diagram of all walls.
%
%
%
The planner chooses the waypoint from the intersection of the local Voronoi diagram and the lookahead circle.
Recall that the \emph{lookahead circle} is a circle of fixed radius centered at the rear axle of the car.
We also assume that the lidar is placed at the middle of the front axle of the car.
We give sufficient conditions such that within the lookahead circle the local and global Voronoi diagrams coincide.
%

Consider the visible subset of the walls, the local Voronoi diagram, and the lookahead circle at some arbitrary time.
Pick a point $p$ on the global Voronoi diagram inside the lookahead circle.
We give sufficient conditions such that the closest wall points to $p$ in the global Voronoi diagram are visible to the lidar. 
Thus, $p$ is also on the local Voronoi diagram.
The sufficient conditions rule out the two possible cases for invisibility of the closest wall points: being out of range, or occluded by a visible point.

Let $R$ be the range of lidar, 
$L$ the distance from lidar to the rear axle,
$\ell$ the lookahead radius,
$m$ the minimum width of the track,
$M$ the maximum width of the track,
and $D$ the minimum distance between lidar and the walls at any point in time.
Then we have the following guarantee:

\begin{theorem}
The local and global Voronoi diagrams coincide within the lookahead circle if
\begin{align}
    R > M, \\
    R > L + \ell + \frac{M}{2}, \label{eq:lidar-range-b} \\
    \textnormal{and,  } D^2 \geq (L+\ell)^2-\frac{m^2}{4}.\label{eq:lidar-wall-distance}
\end{align}
\label{thm:voronoi}
\end{theorem}

Before proving Theorem~\ref{thm:voronoi}, we will prove a lemma about the relationship between lookahead circle, walls of the circuit, and distance between a point on the Voronoi diagram and the walls.

\begin{lemma}
\begin{enumerate}
\item Any point on (local or global) Voronoi diagrams is at most $\frac{M}{2}$ away from its closest walls.
\item Lookahead disk is contained in the circle $C_{L+\ell}$ of radius $L+\ell$ centered at lidar.
\item For any point $p$ on the global Voronoi diagram inside $C_{L+\ell}$, $p$'s closest walls are in lidar's range.
\end{enumerate}
\end{lemma}
\begin{proof}
\begin{enumerate}
    \item 
    Lidar can see its closest walls since $R>M$. Furthermore, a point on the Voronoi diagram is equidistant to its closest walls.
    \item
    The distance from lidar to rear axle is $L$, and lookahead circle is the circle of radius $\ell$ centered at the rear axle.
    \item
    The closest wall points to $p$ are in the circle of radius $L+\ell+\frac{M}{2}$ centered at lidar. By Equation \ref{eq:lidar-range-b}, $p$'s closest wall points are in lidar's range.
\end{enumerate}
\end{proof}

\begin{figure}
\centering
\includegraphics[width=2.5in]{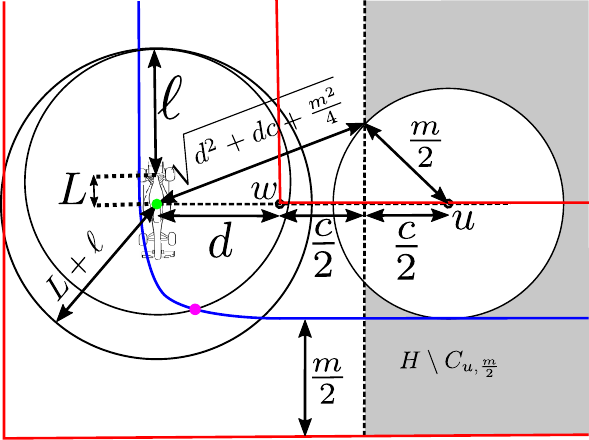}
\caption{Consistency of local and global Voronoi diagrams. The blue curve is the global Voronoi diagram. The green dot is lidar, and the pink dot is the waypoint. The gray area is $H\setminus C_{u, \frac{m}{2}}$. }
\label{fig:voronoi-proof}
\end{figure}

\begin{proof}[Proof of Theorem \ref{thm:voronoi}]
We need to show that for any point $p$ on the global Voronoi diagram, if $p$ is inside the lookahead circle then $p$'s closest walls are visible from lidar, so that $p$ is also on the local Voronoi diagram.
By the lemma, the circle $C_{L+\ell}$ of radius $L+\ell$ centered at lidar contains the lookahead circle, so it is sufficient to assume that $p$ is in $C_{L+\ell}$.
Consider an arbitrary visible wall point $w$ occluding a wall point $u$.
That is, both $w$ and $u$ are on the same ray from lidar but $u$ is further away than $w$.
We show that $p$ is closer to $w$ than $u$.
Let $d$ be the distance of lidar to $w$ and $c$ be the distance of $w$ to $u$ (so the distance of lidar to $u$ is $d+c$).
First note that there is no Voronoi point inside the circle $C_{u, \frac{m}{2}}$ of radius $\frac{m}{2}$ centered at $u$, since $m$ is the minimum width of the track.
Also, the points that are closer to $u$ than $w$ constitute a half-plane $H$ with distance $d+\frac{c}{2}$ from lidar and its boundary being the bisector of $u$ and $w$.
Hence Voronoi points that are closer to $u$ than $w$ are in $H\setminus C_{u, \frac{m}{2}}$ i.e. in $H$ and outside of $C_{u, \frac{m}{2}}$.
See Fig. \ref{fig:voronoi-proof}.
If
$(d+\frac{c}{2})^2 + \frac{m^2}{4} - \frac{c^2}{4} > (L+\ell)^2$
or equivalently
$ d^2+dc > (L+\ell)^2-\frac{m^2}{4} $
then $C_{L+\ell}$ does not intersect $H\setminus C_{u, \frac{m}{2}}$.
The stronger condition 
\begin{equation}
     d^2 \geq (L+\ell)^2-\frac{m^2}{4}
\end{equation}
ensures that a global Voronoi point inside $C_{L+\ell}$ is closer to $w$ than any wall point in $w$'s shadow.
This inequality is guaranteed by Inequality \ref{eq:lidar-wall-distance}, since $D$ is the minimum distance of lidar to walls so $d^2 \geq D^2$.
\end{proof}

Observe that Theorem~\ref{thm:voronoi} gives us some bounds on the size of the lookahead circle. 
The lookahead circle cannot be too large, else, the closest walls to a points in the lookahead circle might be occluded from the lidar.


%% file: sections/6-2-system_model.tex
\subsection{Coupled model of control and plan}
\label{sec:model}



As described in Section~\ref{sec:ppcontroller}, the pure-pursuit controller determines the steering angle $\delta$ according to the waypoint $(g_x, g_y)$ by the planner (given in Equation~\ref{eqn:purepursuit}). 
This waypoint lies on the local Voronoi diagram and always at a distance $\ell$ from the rear-axel. 
When the conditions given in Theorem~\ref{thm:voronoi} are satisfied, the waypoint also lies on the global Voronoi diagram.
Given that the Voronoi diagram is a conjunction of several line segments, without loss of generality, we assume that the waypoint lies on one such line segment.
%

Observe that as the vehicle moves towards the waypoint, the position and orientation of the vehicle changes.
This indeed changes the intersection of the lookahead circle with the Voronoi diagram and as a result, changes the waypoint.
Thus, the evolution of the state of the vehicle (position and orientation) and the waypoint are tightly coupled.
%
In this section, we model the joint behavior of the waypoint along with the vehicle dynamics.

Observe that for pure-pursuit controller, the waypoint $(g_x, g_y)$ is described in the vehicle's coordinate system (i.e., the origin being the center of the rear-axel and the orientation of the vehicle as the x-axis).
We transform this waypoint along the coordinate axis where the Voronoi line segment is the x-axis, the rear wheel has coordinates $(x,y)$ and the orientation of the vehicle $\theta$ is w.r.t the line segment on the Voronoi diagram.
After the coordinate transformation, the orientation $\delta$ in this coordinate system will be
$$ tan(\delta) = \frac{2L}{\ell^2} (-\sqrt{\ell^2-y^2} \cdot sin(\theta)-y \cdot cos(\theta)). $$
Therefore,
\begin{align}
\dot{\theta} & = \frac{\displaystyle v}{\displaystyle L} \frac{\displaystyle 2L}{\displaystyle \ell^2} (-\sqrt{\ell^2-y^2} \cdot sin(\theta)-y \cdot cos(\theta)) \nonumber \\
& = \frac{\displaystyle 2v}{\displaystyle \ell^2} (-\sqrt{\ell^2-y^2} \cdot sin(\theta)-y \cdot cos(\theta))
\end{align}

The closed loop behavior of the vehicle dynamics is therefore given as:
\[
\begin{array}{l}
     \dot{x} = v \cdot cos(\theta) \\
     \dot{y} = v \cdot sin(\theta) \\
     \dot{\theta} = \frac{\displaystyle 2v}{\displaystyle \ell^2} (-\sqrt{\ell^2-y^2} \cdot sin(\theta)-y \cdot cos(\theta))
\end{array}
\]

Notice that this closed loop behavior is accurate as long as the waypoint lies on the same line segment.
When the vehicle makes progress, the lookahead circle would intersect with a different line segment on the voronoi diagram.
As a result, the coordinate system for modeling the evolution of vehicle state and the waypoint changes.
We model this change as a hybrid automata.

\begin{definition} 
A hybrid automata is defined as a tuple $\langle Mod, X, E, Flows \rangle$ where
\begin{itemize}
    \item \textbf{$Modes$} is the set of discrete modes,
    \item \textbf{$X$} is the state space,
    \item \textbf{$E$} is the set of discrete transitions among modes, and
    \item \textbf{$Flows$} describes the evolution of the system in each mode.
\end{itemize}
\end{definition}

Often, $Flows$ are described as a collection of nonlinear differential equations, one for each mode.
Each discrete transition $e = (mode, mode') \in E$ among the modes $mode, mode' \in Modes$ of the hybrid automata has an associated $guard(e)$ condition. The hybrid automata can take the transition only when the state of the trajectory satisfies the guard condition.
Additionally, after taking the discrete transition $e$, the state of the system changes from its current state $x$ to a new state $x'$ defined according to a reset function $x' = reset(e,x)$.

\begin{figure}
\centering
\includegraphics[width=0.8\linewidth]{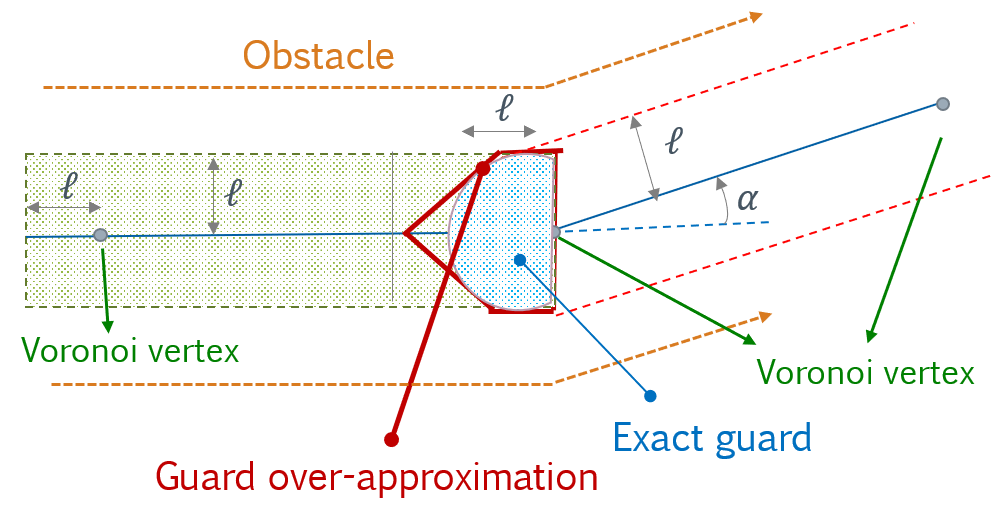}
\caption{Over-approximation of guard.  $l$ is the look ahead distance of the car. The original exact guard is circular and we linearize it as a convex polytope that is defined using linear constraints.}
\label{fig:guard_inv_approx}
\end{figure}

In the case of the vehicle following the waypoint on Voronoi diagram, when a new line segment is encountered by the lookahead circle, the change of basis variables (alignment of the x-axis along the new segment) is the reset function.
Notice that the set of states that take this transition lies on a circle of radius $\ell$ from an end point of the new line segment.
This set is a non-convex set.
Performing analysis on such hybrid automata with non-convex guard conditions is very challenging.
We therefore construct an abstract hybrid automata that allows more behaviors than the original hybrid automata, but is easier to analyze.
More specifically, we allow the hybrid automata to non-deterministically take a transition whenever the vehicle enters the convex overapproximation of the original guard condition.
Given that we strictly increase the set of possible states that can take the discrete transitions, it is easy to observe that the abstract hybrid automata includes all the behaviors of the original hybrid automata.
Illustration of the abstraction of the guard condition for hybrid automata modeling the behavior of the vehicle is given in Figure~\ref{fig:guard_inv_approx}.

Given a circuit, we construct the hybrid automata model of the vehicle moving in the lap using our Voronoi planning and pure pursuit control algorithm.
In the next section, we present the details of the techniques employed to prove the safety and progress properties of the vehicle.

%% file: sections/6-3-reachable-set.tex
\subsection{Reachable Set Computation}

\begin{definition}

Given a hybrid system $H$ modeled as an hybrid automaton and an initial set of states $\Theta$, an execution of $H$ is a sequence of trajectories and transitions $\xi_0 e_1 \xi_1 e_2 \ldots $ such that
(i) the first state of $\xi_0$ denoted as $q_0$ is in the initial set, i.e., $q_0 = (mode_0, x_0)\in \Theta$,
(ii) each trajectory $\xi_i$ is the solution of the differential equation $Flows_{i}$ of the corresponding mode $mode_i$, 
(iii) the state of the trajectory before each discrete transition $e_i = (mode_i, mode_{i+1})$ satisfies $guard(e_i)$,
(iv) and the state of the trajectory after taking the transition $e_i$ changes to $q_{i+1} = (mode_{i+1}, x_{i+1})$ where $x_{i+1} = reset (e_i, x_i)$.
\end{definition}
The set of states encountered by all executions that conform to the above semantics is called the \emph{reachable set} and is denoted as $Reach_{[H, \Theta]}$. Bounded-time variant of these executions and the reachable set defined over the time bound $T$ is denoted as $Reach_{[H, \Theta]}^T$. We drop $H$ and/or $\Theta$ from $Reach_{[H, \Theta]}^T$ whenever it is clear from the context, and abuse the term \emph{trajectory} to denote the hybrid system execution as well as the solution of the differential equation of a mode.

We say the reachable set computation has a \emph{fixed point at time $t$} when there exists $t \leq T$ such that $Reach^t = Reach^{t+1}$.
Since it is computationally hard to compute the exact reachable set for most system classes including hybrid systems ~\cite{ALUR19953,10.1007/3-540-46430-1_6}, a verification engine is typically used to compute an overapproximation of the reachable set denoted as $\tilde{Reach^T}$. Given the sequence of reachable set computed at discrete time instances, the $i^{th}$ element in the is denoted as $Reach^T[i]$.

\vspace{0.2cm}
\begin{definition}
\label{def:hybridSafe}
A hybrid system $H$ with initial set $\Theta$, time bound $T$, and unsafe set $U$ is said to be safe with respect to its executions if all trajectories starting from $\Theta$ for bounded time $T$ are safe i.e., $\tilde{Reach^T} \cap U = \emptyset$.
\end{definition}

As the reachable set overapproximation $\tilde{Reach^T}$ includes more behaviors than the exact reachable set, its safety w.r.t. $U$ proves the safety of the exact reachable set w.r.t. $U$. However, the safety result is inconclusive when $\tilde{Reach^T} \cap U \neq  \emptyset$. We next discuss the fixed point based computation of the reachable set for a given initial set $\Theta$. 

\begin{itemize}[leftmargin=*]
\item \textbf{Computing a Fixed Point :} Recall that the vehicle motion for a given circuit is modeled as the hybrid automata where each mode is associated with an edge from the global Voronoi diagram.
We denote the lap as one full pass of the given circuit which is the sequence of edges from the global Voronoi diagram such that an end point of the last segment coincides with an initial point of the first segment. Informally, the number of laps represents the number of passes of the given circuit performed by the vehicle.
%

As the vehicle completes one lap while following Voronoi edges by switching modes of the hybrid automaton, it may not arrive back at the exact system state it initially started from. As a consequence, the set of states reachable in a mode during next lap is not necessarily same as the set of states reachable in this mode during previous lap(s). Whereas, a fixed point at some time $t$ is an evidence that the set of reachable states beyond time $t$ is time-invariant. 

Computing a fixed point for the reachable set serves three important purposes - (i) arriving at the same mode(s) after completing a lap underscores \emph{progress}, (ii) safety of the fixed point at $t$ ensures \emph{safety} of the reachable set at all times beyond $t$, (iii) it makes the analysis \emph{efficient} because one can save a significant amount of computational resources by not requiring to compute the reachable set after $t$ for a much larger initial set as explained next.

%
%

\item \textbf{Computing additional reachable sets :} The error bound of reachable set overapproximation is proportional to the size of the initial set $\Theta$, i.e., larger the initial set, higher is the overapproximation error. 
Whereas, computing the reachable set for each state in the infinite state system as ours is practically impossible. 
The standard approach for handling this trade-off is to refine the large initial set into smaller subsets and perform reachable set computation on each subset.

A fixed point can assist in accelerating the reachable set computation for a large set $\Theta$. For any two given sets $\theta, \theta' \in \Theta$, it may not be necessary to compute the fixed point of $Reach_{\theta'}$ once the fixed point  $Reach_{\theta}^t$ at time $t$ is obtained. 
While computing $Reach_{\theta'}$, we iteratively check whether $Reach_{\theta'}^i[i] \subseteq Reach_{\theta}^t[i], i \leq t$, and halt the computation as soon as the containment check returns true. 
Generalizing this approach to more than two sets would mean computing the fixed point of one set and perform the containment check for the rest w.r.t. the fixed point.

\item \textbf{Refinement techniques :} One way to computing the   reachable set for a large set $\Theta$ is to \emph{manually} obtain the fixed point candidate $\theta \in \Theta$, and find the partitions $\theta' \in \Theta, \theta \cup \theta' = \Theta$ such that each partition has its reachable set at some $i \leq t$  contained in the fixed point. 
An efficient strategy is to \emph{automatically} obtain both the fixed point candidate and other partitions. If the overapproximation error is too high for the given set $\Theta$, it is automatically refined into multiple sets. 
One of these subsets is picked for the fixed point candidacy. If the overapproximation error for the candidate  set is still high, it is further divided into parts until a candidate with its fixed point at some $t$ is found. The rest of $\Theta$ is automatically partitioned depending on whether their reachable sets are contained in the fixed point at $i \leq t$ or not.
\end{itemize}


%% file: sections/8-0-evaluation.tex
\section{Evaluation}
In this section, we evaluate our fixed-point based reachable set computation on five different tracks with different characteristics. 
For each track given as the sequence of global Voronoi diagram edges, a new hybrid automaton model for the evolution of vehicle along this track is generated. Typically, a hybrid automata has as many discrete modes as the number of Voronoi edges. 
The dynamics in each mode of the  automaton is the closed loop behavior of the vehicle dynamics and the way point as described in Section~\ref{sec:model}.
Then, we employ a non-linear hybrid systems verification tool to computed the fixed point of the reachable set for each track.
%
The computation of a fixed point in each track underscores both safety and progress of the vehicle.
In addition, we tested the planning and control algorithm both in simulation and open source F1Tenth platform.\footnote{\url{https://tinyurl.com/ry5xhza}}

\subimport{}{8-1-verification.tex}

%% file: sections/8-1-verification.tex


\begin{figure*}
    \centering
    \subfigure[for small initial set]{\label{fig:track1_voronoi-cora}\includegraphics[width=3.2in]{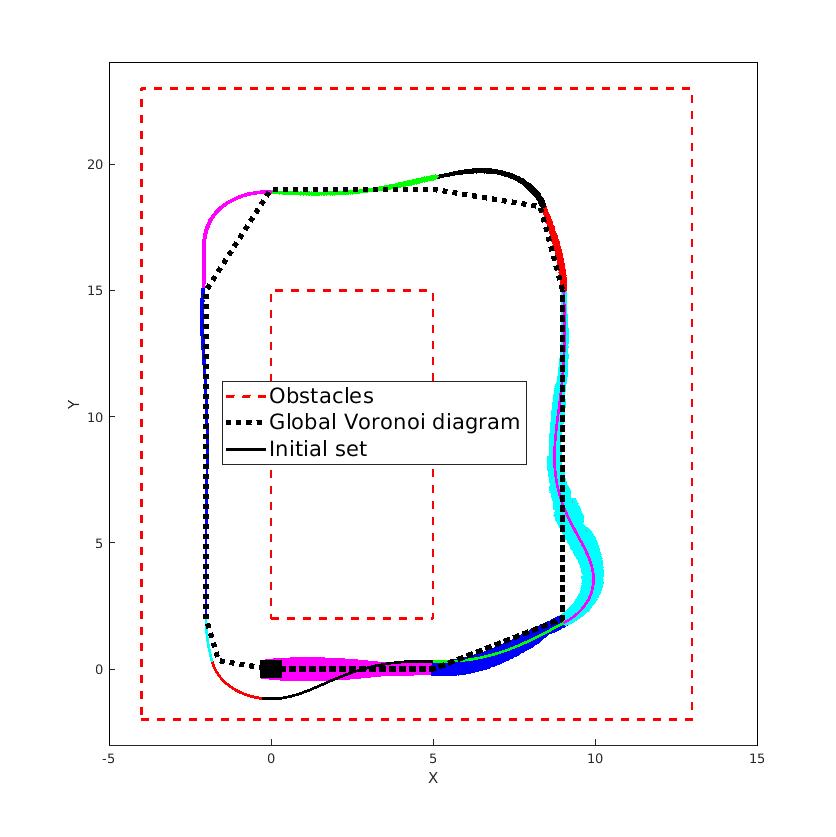}}
    \subfigure[for a large initial  set]{\label{fig:track1_cora-abstraction}\includegraphics[width=3.2in]{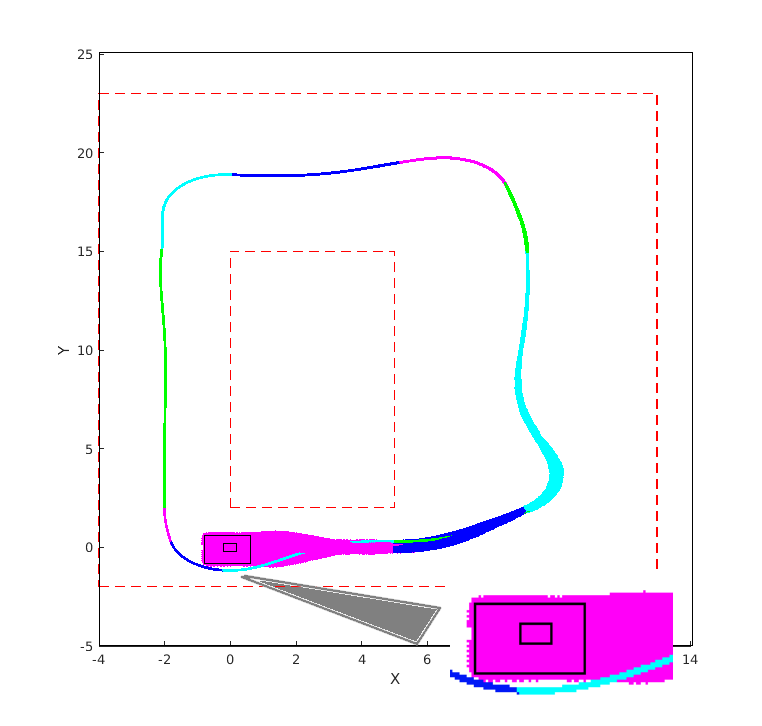}}
    \caption{\textbf{Reachable set computation for Track-1}. In Fig.~\ref{fig:track1_voronoi-cora}, the fixed point is obtained in mode 3 in the second lap. The filled black rectangle depicts the initial set. In Fig.~\ref{fig:track1_cora-abstraction}, the smaller rectangle corresponds to the fixed point partition and larger one is the originally given initial set.}
\label{fig:eval_track1}
\end{figure*}


We experiment with multiple safety verification platforms such as CORA~\cite{Althoff2018b}, Flow*~\cite{10.1007/978-3-642-39799-8_18}, and C2E2~\cite{Duggirala.2015} for non-linear hybrid systems. Different platforms use different symbolic representations for the reachable set. For instance, Flow* uses Taylor models, C2E2 uses Jacobian matrix and discrepancy functions, and CORA primarily makes use of zonotopes for representing the reachable sets. As their utility is application specific,  we observed that overapproximation error in both C2E2 and Flow* is very high for our case studies. One possible reason could be that Taylor model approximation in Flow* and discrepancy function computed in C2E2 are too conservative. We illustrate the  reachable set accuracy across Flow* and CORA on a small system in Section~\ref{subsec:appendix-reach-set-accuracy} in the Appendix.



%


In the safety verification of our dynamical model, the obstacles or walls constitute the unsafe set i.e., the objective is to verify that the reachable set does not overlap with the walls. 
%
%
The initial set considered during reachable set computation in track-1 is $x \in [-0.3, 0.3]$, $y \in [-0.3, 0.3]$ and $\theta \in [-0.2, 0.2]$.
The Voronoi diagram and the corresponding reachable set for track-1 are shown in Fig.~\ref{fig:track1_voronoi-cora}. Based on the track width, the turns in this track encompass 4 different transitions - wide to narrow, wide to wide, narrow to narrow, and narrow to wide. 
Also observe that the fixed point of the reachable set computation is obtained in \emph{mode 3} in lap 2. The values of state variables in \emph{mode 3} across lap-1 and lap-2 are depicted in Fig.~\ref{fig:track1-fixed-point-loc3} that establishes our fixed point claim. The reachable set computation results for other tracks are illustrated in Fig.~\ref{fig:eval_track2-track3-track4-track5} in the Appendix. CORA takes roughly a minute to compute the reachable set for each track.


%
%
An observation from the safety verification results is that if the set of initial conditions is large (larger uncertainty of vehicle's initial orientation and position with respect to the map), CORA fails to compute the reachable set over-approximation. 
As suggested earlier, this can potentially be mitigated by fixed-point based partitioning of the large initial set and verifying the safety for each partition.

We have implemented the automatic partitioning algorithm and evaluated the approach on track-1 for the initial set $x \in [-0.8, 0.6]$, $y \in [-0.8, 0.6]$ and $\theta \in [-0.2, 0.2]$. CORA fails to compute the reachable set for this initial set. 
Our technique keeps partitioning the set until it finds a  partition $x \in [-0.2, 0.2]$, $y \in [-0.2, 0.2]$ and $\theta \in [-0.2, 0.2]$ having a fixed point. (We do not partition $\theta$ in this case).
Once the fixed point is found, the rest of the initial set is partitioned into intervals of the given width and reachable set is computed for each partition. In addition, at each step, reachable set containment w.r.t. the fixed point is conducted to avoid redundant computation. Fig.~\ref{fig:track1_cora-abstraction} demonstrates the safe reachable set successfully computed by our approach  for the given set.

We also performed reachable set computation using CORA over tracks where the vehicle had to take a steep turn (turns of more than 90 degrees). For such instances, CORA failed to compute an overapproximation of the reachable set that is sufficient to establish the safety even for small initial sets. This shows that there is room for improvement in the current reachable set computation methods.


%% file: sections/9-conclusion.tex
\section{Conclusion and Future Work}
We  have illustrated the reachability based verification results of a reactive planning and control autonomous vehicle that uses Voronoi diagrams for planning and pure-pursuit controller for navigation. 
To the best of authors' knowledge, this is the first work that considers all three aspects of planning, control, and the dynamically changing waypoint for prove the safety specification.
We also demonstrated reactive planning and control technique in various tracks in simulation, and on two physical tracks using a scaled down version of autonomous vehicle. 
%
We believe that the coupled effect of planning and control requires further investigation.
In future, we intend to extend this to prove and demonstrate the safety of autonomous vehicles in presence of dynamic obstacles.
%

%% file: sections/appendix.tex
\section{Appendix}

\subsection{Reachable set accuracy}
\label{subsec:appendix-reach-set-accuracy}
\begin{figure}
    \centering
    \subfigure[Flow*]{\label{fig:reachSet1_flow}\includegraphics[width=3.5in]{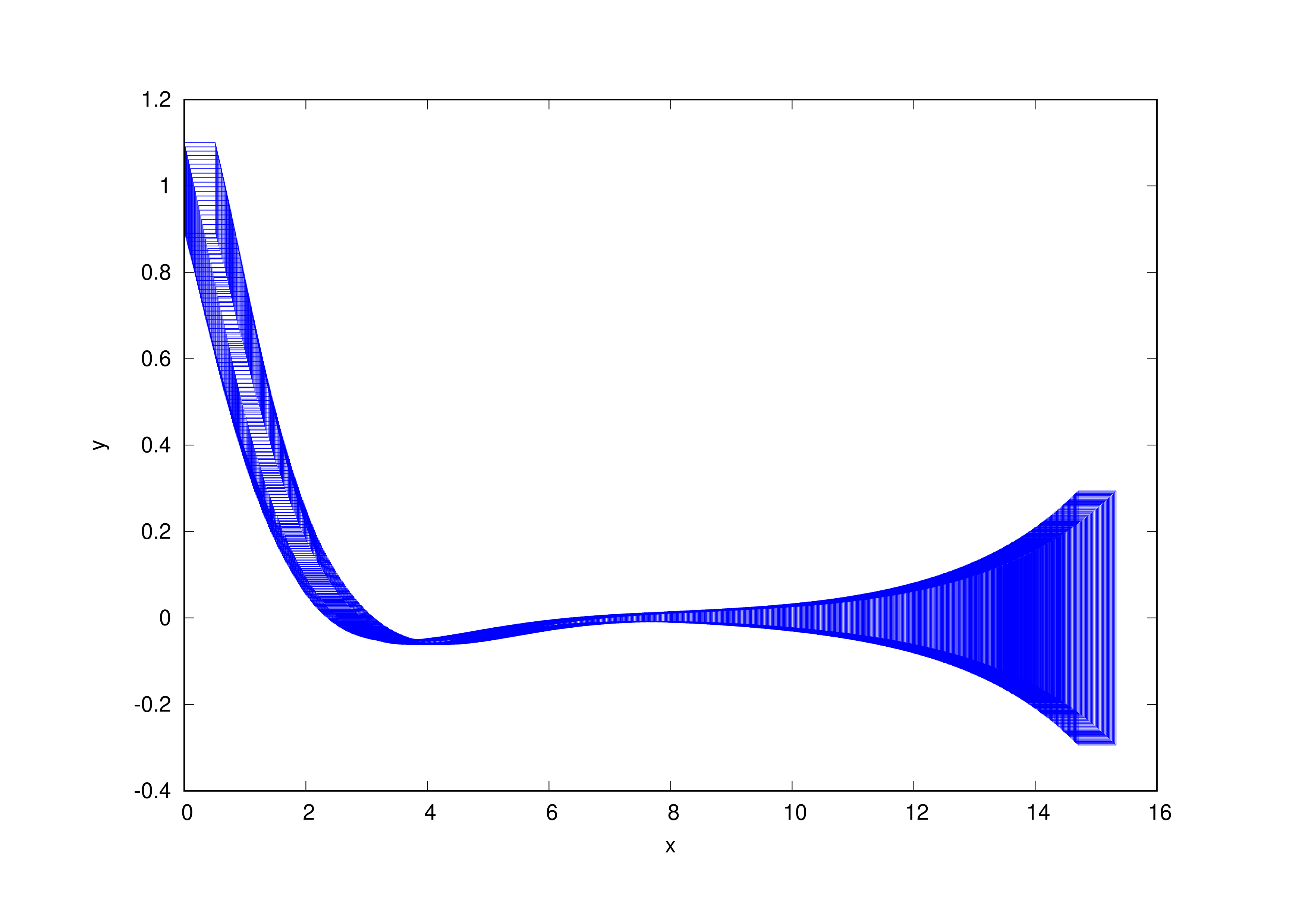}}
    \subfigure[CORA]{\label{fig:reachSet1_cora}\includegraphics[width=\linewidth]{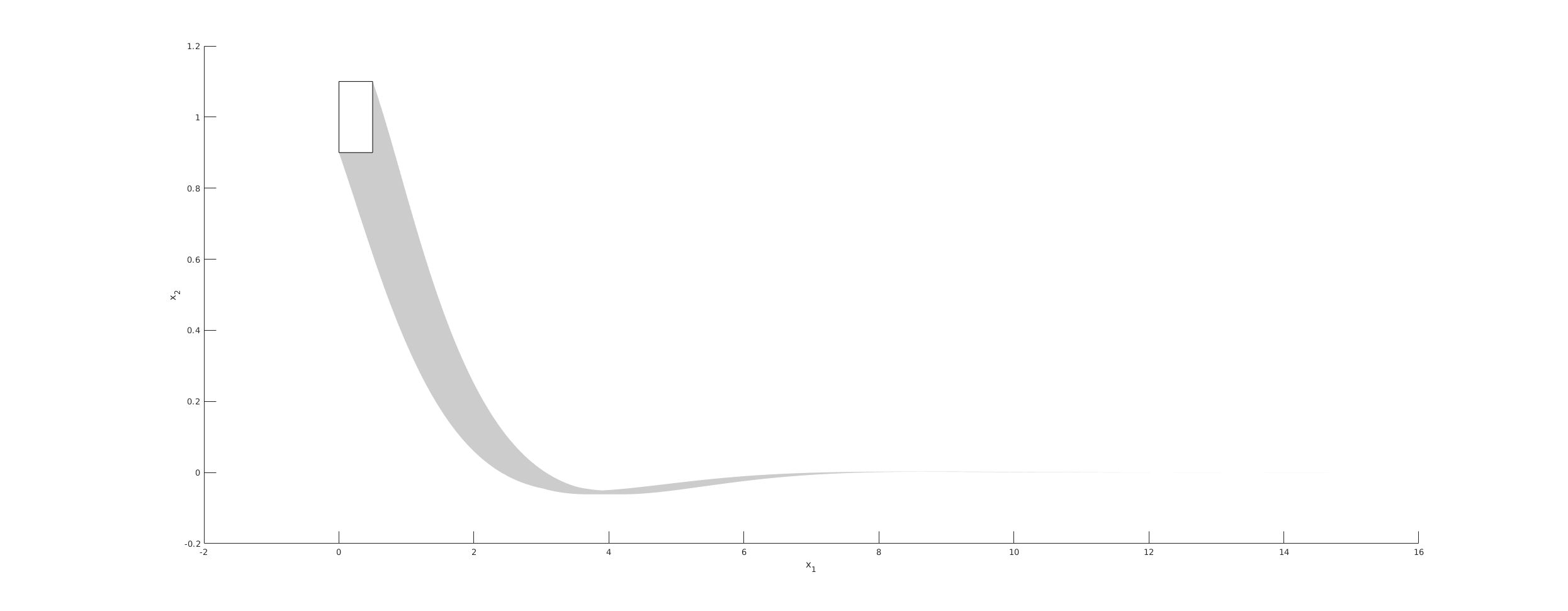}}
    \caption{Reachable sets computed in Flow* and CORA with time step 0.02 sec and time bound 15 sec for a set of initial states. The vehicle follows a vertical path downwards before making  a left turn.}
\label{fig:reachSets}
\end{figure}

Suppose the track is defined in a stationary $(x, y)$ frame, $\theta$ is the vehicle's orientation and 
$\ell$ is the look ahead distance. 
%
As a part of experimentation, we perform the reachable set computation in Flow* and CORA by modeling a single turn on the track as the hybrid automata. 
The initial set is given as intervals over state variables i.e., $x \in [0.0, 0.5]$, $y \in [0.9, 1.1]$ and $\theta \in [-0.5, -0.5]$. 
The reachable sets computed in both tools are shown in figures~\ref{fig:reachSet1_flow} and ~\ref{fig:reachSet1_cora}. 
The divergent behavior of the reachable set in Flow* is seemingly due to error compounded over time because of coarse approximation. 
The figures also illustrate that the vehicle while turning swings to some extent before merging back on to the track.

\begin{figure*}
    \centering
    \subfigure[Track 1-X]{\label{fig:track1-loc3-x}\includegraphics[width=3.8in]{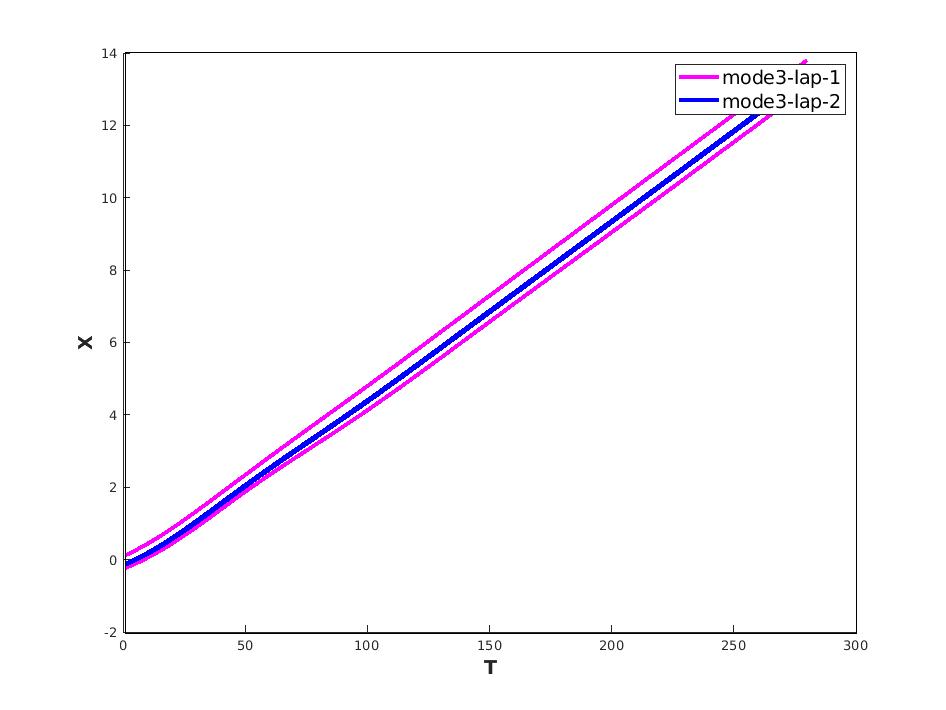}}
    \subfigure[Track 1-Y]{\label{fig:track1-loc3-y}\includegraphics[width=3.8in]{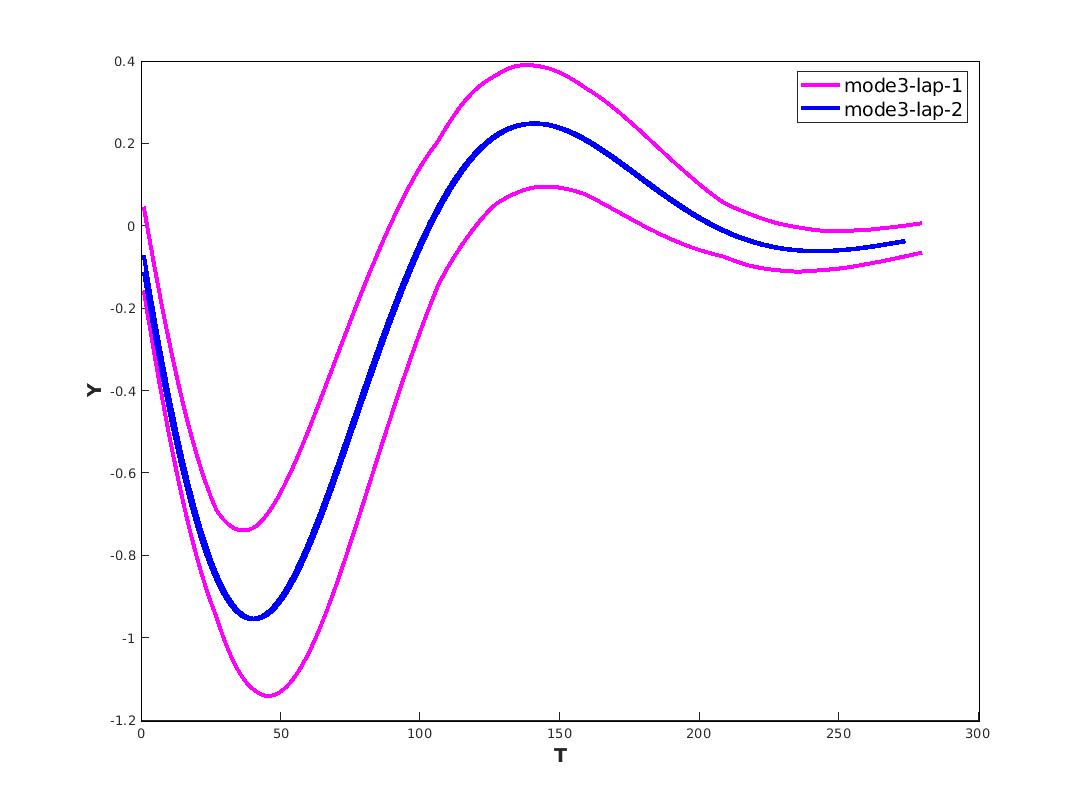}}
    \subfigure[Track 1-$\theta$]{\label{fig:track1-loc3-theta}\includegraphics[width=3.8in]{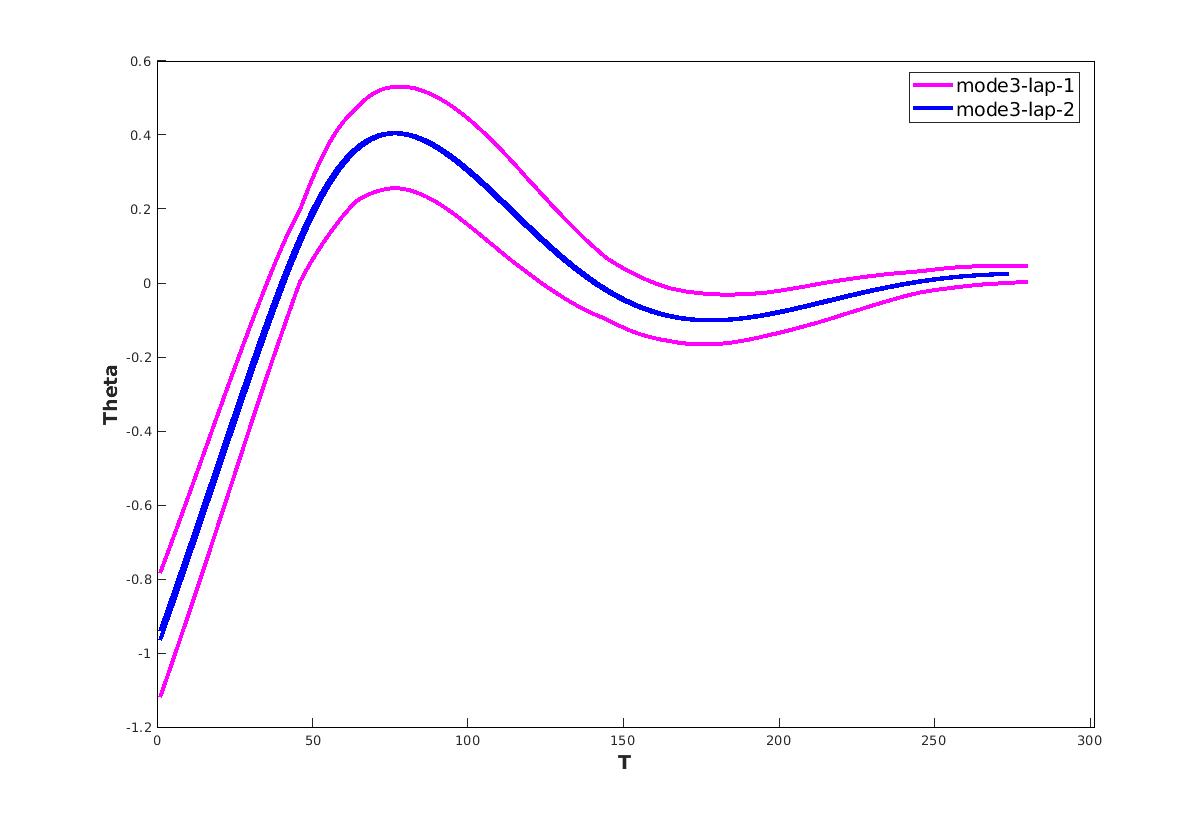}}
    \caption{\textbf{Fixed point illustration using state variables in Track-1}}
\label{fig:track1-fixed-point-loc3}
\end{figure*}


\begin{figure*}
    \centering
    \subfigure[Track 2]{\label{fig:track2-voronoi-cora}\includegraphics[width=3.5in]{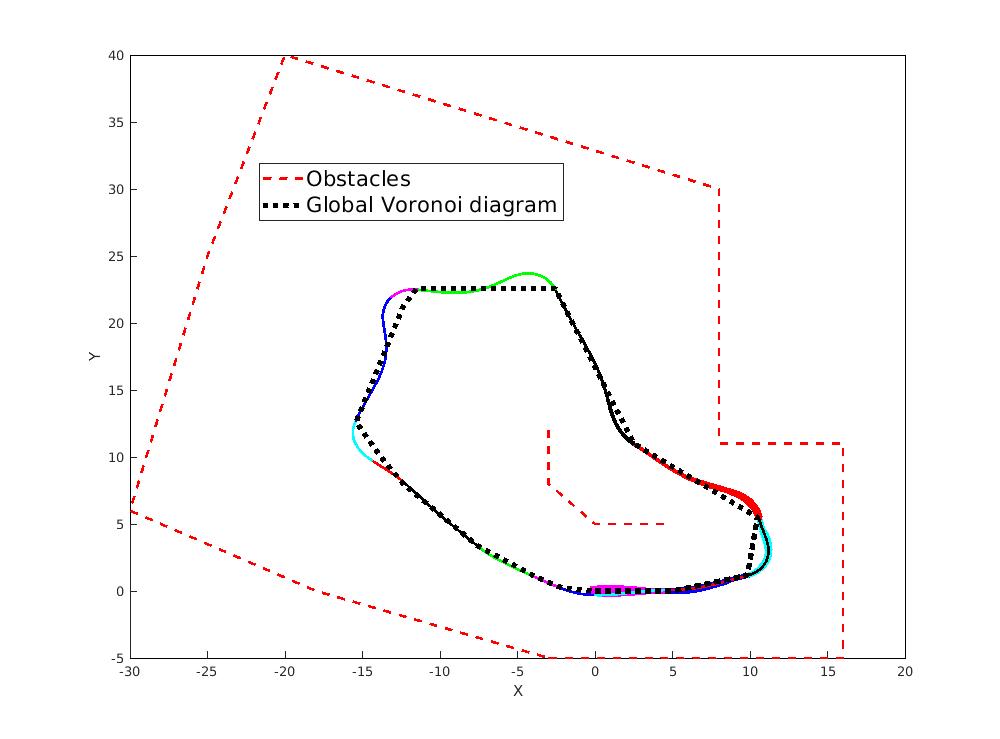}}
    \subfigure[Track 3]{\label{fig:track3-voronoi-cora}\includegraphics[width=3.5in]{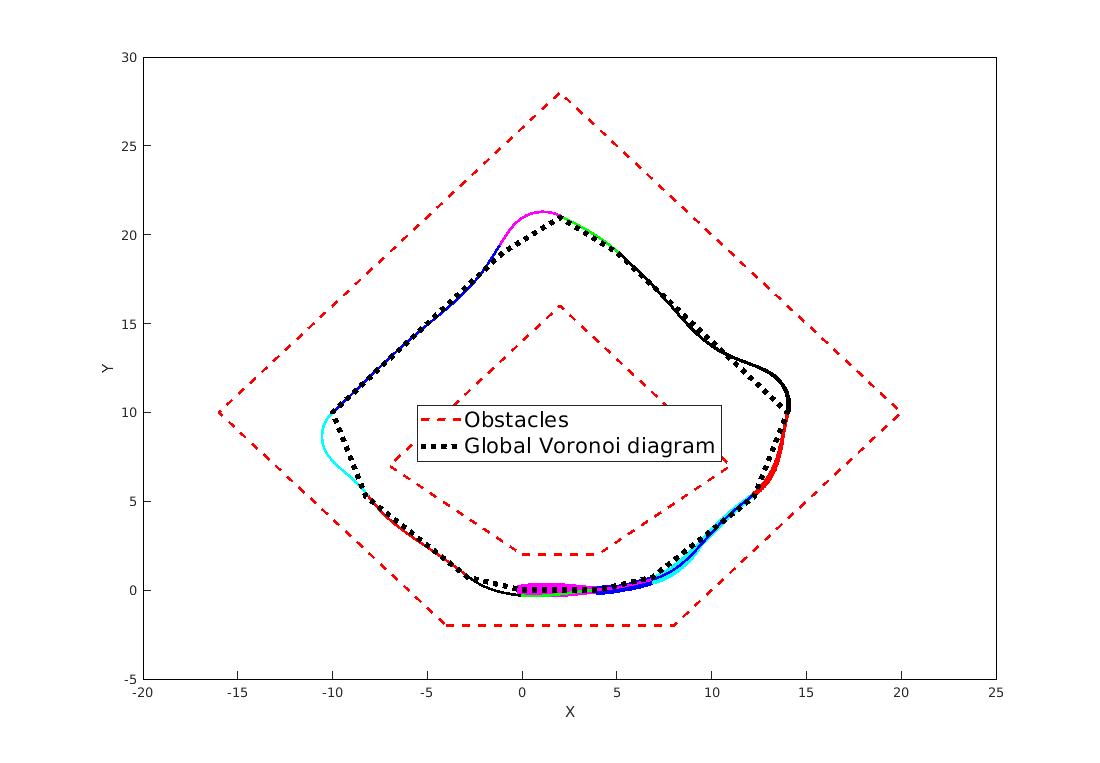}}
    \subfigure[Track 4]{\label{fig:track4-voronoi-cora}\includegraphics[width=3.5in]{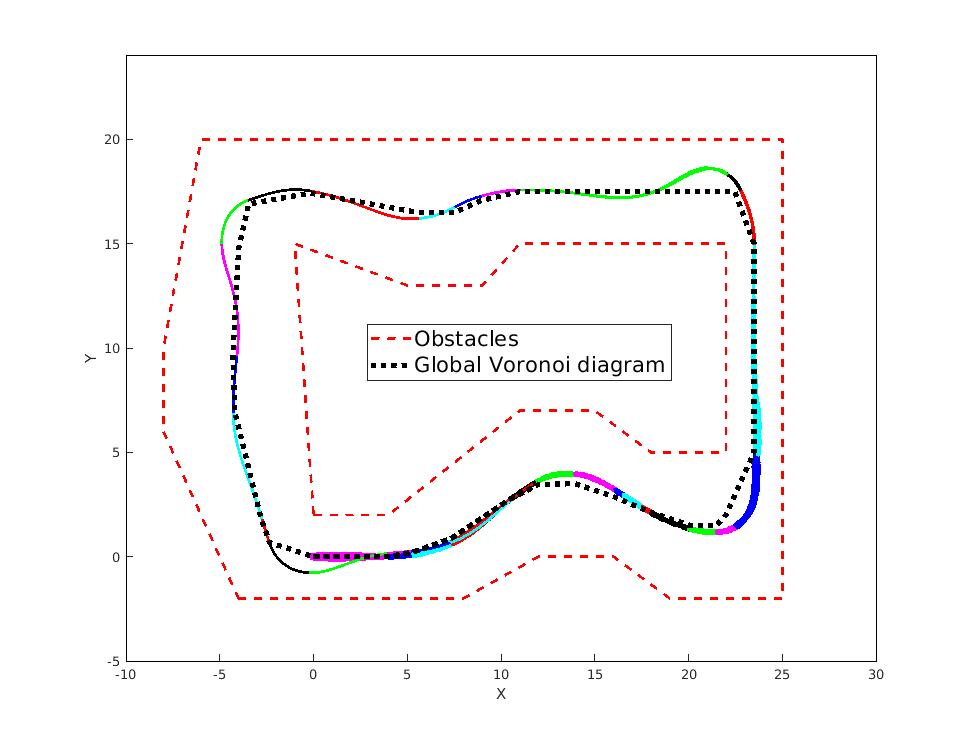}}
    \subfigure[Track 5]{\label{fig:track5-voronoi-cora}\includegraphics[width=3.5in]{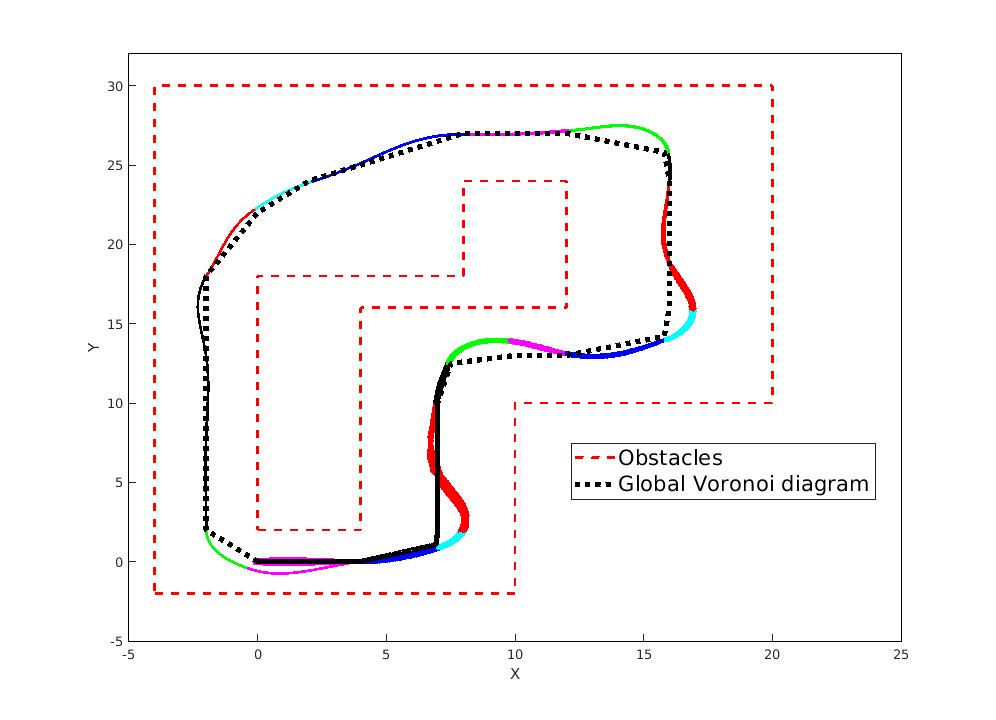}}
    \caption{Safety verification of the plan in multiple other tracks. The track-wise initial sets are $\Theta_{2} = [[-0.25, 0.25][-0.25, 0.25][-0.15,0.15]]$, $\Theta_{3} = [[-0.2, 0.2][-0.2, 0.2][-0.2,0.2]]$, $\Theta_{4} = [[-0.12, 0.12][-0.12, 0.12][-0.12,0.12]]$, and $\Theta_{5} = [[-0.12, 0.12][-0.12, 0.12][-0.12,0.12]]$}
\label{fig:eval_track2-track3-track4-track5}
\end{figure*}


%% file: main.bbl
\begin{thebibliography}{54}
\providecommand{\natexlab}[1]{#1}
\providecommand{\url}[1]{\texttt{#1}}
\expandafter\ifx\csname urlstyle\endcsname\relax
  \providecommand{\doi}[1]{doi: #1}\else
  \providecommand{\doi}{doi: \begingroup \urlstyle{rm}\Url}\fi

\bibitem[f1t()]{f1tenth}
F1tenth website.
\newblock URL \url{https://f1tenth.org/}.

\bibitem[{Althoff} and {Dolan}(2014)]{AlthoffDolan}
M.~{Althoff} and J.~M. {Dolan}.
\newblock Online verification of automated road vehicles using reachability
  analysis.
\newblock \emph{IEEE Transactions on Robotics}, 2014.

\bibitem[Althoff et~al.(2018)Althoff, Grebenyuk, and Kochdumper]{Althoff2018b}
M.~Althoff, D.~Grebenyuk, and N.~Kochdumper.
\newblock Implementation of taylor models in cora 2018.
\newblock In \emph{Proc. of the 5th International Workshop on Applied
  Verification for Continuous and Hybrid Systems}, pages 145--173, 2018.

\bibitem[Alur et~al.(1995)Alur, Courcoubetis, Halbwachs, Henzinger, Ho,
  Nicollin, Olivero, Sifakis, and Yovine]{ALUR19953}
R.~Alur, C.~Courcoubetis, N.~Halbwachs, T.A. Henzinger, P.-H. Ho, X.~Nicollin,
  A.~Olivero, J.~Sifakis, and S.~Yovine.
\newblock The algorithmic analysis of hybrid systems.
\newblock \emph{Theoretical Computer Science}, 138\penalty0 (1):\penalty0
  3--34, 1995.
\newblock ISSN 0304-3975.
\newblock \doi{https://doi.org/10.1016/0304-3975(94)00202-T}.
\newblock URL
  \url{https://www.sciencedirect.com/science/article/pii/030439759400202T}.
\newblock Hybrid Systems.

\bibitem[Amidi and Thorpe(1991)]{Amidi.1991}
Omead Amidi and Chuck~E Thorpe.
\newblock Integrated mobile robot control.
\newblock In \emph{Mobile Robots V}, volume 1388, pages 504--523. International
  Society for Optics and Photonics, 1991.

\bibitem[Aoude et~al.(2010)Aoude, Luders, Levine, and How]{aoude2010threat}
Georges~S Aoude, Brandon~D Luders, Daniel~S Levine, and Jonathan~P How.
\newblock Threat-aware path planning in uncertain urban environments.
\newblock In \emph{2010 IEEE/RSJ International Conference on Intelligent Robots
  and Systems}, pages 6058--6063. IEEE, 2010.

\bibitem[Asarin et~al.(2000)Asarin, Bournez, Dang, and
  Maler]{10.1007/3-540-46430-1_6}
Eugene Asarin, Olivier Bournez, Thao Dang, and Oded Maler.
\newblock Approximate reachability analysis of piecewise-linear dynamical
  systems.
\newblock In Nancy Lynch and Bruce~H. Krogh, editors, \emph{Hybrid Systems:
  Computation and Control}, pages 20--31, Berlin, Heidelberg, 2000. Springer
  Berlin Heidelberg.
\newblock ISBN 978-3-540-46430-3.

\bibitem[Belkhouche(2009)]{belkhouche2009reactive}
Fethi Belkhouche.
\newblock Reactive path planning in a dynamic environment.
\newblock \emph{IEEE Transactions on Robotics}, 25\penalty0 (4):\penalty0
  902--911, 2009.

\bibitem[Bruce and Veloso(2002)]{bruce2002real}
James Bruce and Manuela~M Veloso.
\newblock Real-time randomized path planning for robot navigation.
\newblock In \emph{Robot Soccer World Cup}, pages 288--295. Springer, 2002.

\bibitem[Chen et~al.(2013)Chen, {\'A}brah{\'a}m, and
  Sankaranarayanan]{10.1007/978-3-642-39799-8_18}
Xin Chen, Erika {\'A}brah{\'a}m, and Sriram Sankaranarayanan.
\newblock Flow*: An analyzer for non-linear hybrid systems.
\newblock In Natasha Sharygina and Helmut Veith, editors, \emph{Computer Aided
  Verification}, pages 258--263, Berlin, Heidelberg, 2013. Springer Berlin
  Heidelberg.
\newblock ISBN 978-3-642-39799-8.

\bibitem[Das et~al.(2011)Das, Mukherjee, Wang, Salas, Bedi, and
  Waslander]{Das.2011}
A~Das, P~Mukherjee, C~Wang, G~Salas, S~Bedi, and S~Waslander.
\newblock Graph based path planning in unknown environments using voronoi
  diagrams.
\newblock In \emph{Canadian Congress of Applied Mechanics, Proceedings of the
  2011}. Vancouver, BC, 2011.

\bibitem[De~Luca et~al.(1998)De~Luca, Oriolo, and Samson]{de1998feedback}
Alessandro De~Luca, Giuseppe Oriolo, and Claude Samson.
\newblock Feedback control of a nonholonomic car-like robot.
\newblock In \emph{Robot motion planning and control}, pages 171--253.
  Springer, 1998.

\bibitem[Dolgov et~al.(2010)Dolgov, Thrun, Montemerlo, and Diebel]{Dolgov.2010}
Dmitri Dolgov, Sebastian Thrun, Michael Montemerlo, and James Diebel.
\newblock Path planning for autonomous vehicles in unknown semi-structured
  environments.
\newblock \emph{The International Journal of Robotics Research}, 29\penalty0
  (5):\penalty0 485--501, 2010.

\bibitem[Duggirala et~al.(2015)Duggirala, Mitra, Viswanathan, and
  Potok]{Duggirala.2015}
Parasara~Sridhar Duggirala, Sayan Mitra, Mahesh Viswanathan, and Matthew Potok.
\newblock C2e2: A verification tool for stateflow models.
\newblock In \emph{International Conference on Tools and Algorithms for the
  Construction and Analysis of Systems}, pages 68--82. Springer, 2015.

\bibitem[Dumonteil et~al.(2015)Dumonteil, Manfredi, Devy, Confetti, and
  Sidobre]{dumonteil2015reactive}
Gautier Dumonteil, Guido Manfredi, Michel Devy, Ambroise Confetti, and Daniel
  Sidobre.
\newblock Reactive planning on a collaborative robot for industrial
  applications.
\newblock In \emph{2015 12th International Conference on Informatics in
  Control, Automation and Robotics (ICINCO)}, volume~2, pages 450--457. IEEE,
  2015.

\bibitem[Fainekos et~al.(2009)Fainekos, Girard, Kress-Gazit, and
  Pappas]{fainekos2009temporal}
Georgios~E Fainekos, Antoine Girard, Hadas Kress-Gazit, and George~J Pappas.
\newblock Temporal logic motion planning for dynamic robots.
\newblock \emph{Automatica}, 45\penalty0 (2):\penalty0 343--352, 2009.

\bibitem[Falcone et~al.(2007)Falcone, Borrelli, Asgari, Tseng, and
  Hrovat]{falcone2007predictive}
Paolo Falcone, Francesco Borrelli, Jahan Asgari, Hongtei~Eric Tseng, and Davor
  Hrovat.
\newblock Predictive active steering control for autonomous vehicle systems.
\newblock \emph{IEEE Transactions on control systems technology}, 15\penalty0
  (3):\penalty0 566--580, 2007.

\bibitem[Falcone et~al.(2008)Falcone, Borrelli, Tseng, Asgari, and
  Hrovat]{falcone2008linear}
Paolo Falcone, Francesco Borrelli, H~Eric Tseng, Jahan Asgari, and Davor
  Hrovat.
\newblock Linear time-varying model predictive control and its application to
  active steering systems: Stability analysis and experimental validation.
\newblock \emph{International Journal of Robust and Nonlinear Control:
  IFAC-Affiliated Journal}, 18\penalty0 (8):\penalty0 862--875, 2008.

\bibitem[Georgeff and Lansky()]{georgeff1987reactive}
Michael~P Georgeff and Amy~L Lansky.
\newblock Reactive reasoning and planning.

\bibitem[Hoffmann et~al.(2007)Hoffmann, Tomlin, Montemerlo, and
  Thrun]{hoffmann2007autonomous}
Gabriel~M Hoffmann, Claire~J Tomlin, Michael Montemerlo, and Sebastian Thrun.
\newblock Autonomous automobile trajectory tracking for off-road driving:
  Controller design, experimental validation and racing.
\newblock In \emph{2007 American Control Conference}, pages 2296--2301. IEEE,
  2007.

\bibitem[Kant and Zucker(1986)]{kant1986toward}
Kamal Kant and Steven~W Zucker.
\newblock Toward efficient trajectory planning: The path-velocity
  decomposition.
\newblock \emph{The international journal of robotics research}, 5\penalty0
  (3):\penalty0 72--89, 1986.

\bibitem[Karaman and Frazzoli(2011)]{karaman2011sampling}
Sertac Karaman and Emilio Frazzoli.
\newblock Sampling-based algorithms for optimal motion planning.
\newblock \emph{The international journal of robotics research}, 30\penalty0
  (7):\penalty0 846--894, 2011.

\bibitem[Kavraki et~al.(1996)Kavraki, Svestka, Latombe, and
  Overmars]{kavraki1996probabilistic}
Lydia~E Kavraki, Petr Svestka, J-C Latombe, and Mark~H Overmars.
\newblock Probabilistic roadmaps for path planning in high-dimensional
  configuration spaces.
\newblock \emph{IEEE transactions on Robotics and Automation}, 12\penalty0
  (4):\penalty0 566--580, 1996.

\bibitem[{Kloetzer} and {Belta}(2010)]{Kloetzertemp}
M.~{Kloetzer} and C.~{Belta}.
\newblock Automatic deployment of distributed teams of robots from temporal
  logic motion specifications.
\newblock \emph{IEEE Transactions on Robotics}, 2010.

\bibitem[Kress-Gazit et~al.(2009)Kress-Gazit, Fainekos, and
  Pappas]{kress2009temporal}
Hadas Kress-Gazit, Georgios~E Fainekos, and George~J Pappas.
\newblock Temporal-logic-based reactive mission and motion planning.
\newblock \emph{IEEE transactions on robotics}, 25\penalty0 (6):\penalty0
  1370--1381, 2009.

\bibitem[LaValle(1998)]{lavalle1998rapidly}
Steven~M LaValle.
\newblock Rapidly-exploring random trees: A new tool for path planning.
\newblock 1998.

\bibitem[Levinson et~al.(2011)Levinson, Askeland, Becker, Dolson, Held, Kammel,
  Kolter, Langer, Pink, Pratt, et~al.]{levinson2011towards}
Jesse Levinson, Jake Askeland, Jan Becker, Jennifer Dolson, David Held, Soeren
  Kammel, J~Zico Kolter, Dirk Langer, Oliver Pink, Vaughan Pratt, et~al.
\newblock Towards fully autonomous driving: Systems and algorithms.
\newblock In \emph{2011 IEEE Intelligent Vehicles Symposium (IV)}, pages
  163--168. IEEE, 2011.

\bibitem[Loos et~al.(2011)Loos, Platzer, and Nistor]{ACCVerified}
Sarah~M. Loos, Andr{\'e} Platzer, and Ligia Nistor.
\newblock Adaptive cruise control: Hybrid, distributed, and now formally
  verified.
\newblock In Michael Butler and Wolfram Schulte, editors, \emph{FM 2011: Formal
  Methods}, 2011.

\bibitem[{Lygeros} et~al.(1998){Lygeros}, {Godbole}, and
  {Sastry}]{LygerosSastry}
J.~{Lygeros}, D.~N. {Godbole}, and S.~{Sastry}.
\newblock Verified hybrid controllers for automated vehicles.
\newblock \emph{IEEE Transactions on Automatic Control}, 1998.

\bibitem[Miller et~al.(2008)Miller, Campbell, Huttenlocher, Kline, Nathan,
  Lupashin, Catlin, Schimpf, Moran, Zych, et~al.]{miller2008team}
Isaac Miller, Mark Campbell, Dan Huttenlocher, Frank-Robert Kline, Aaron
  Nathan, Sergei Lupashin, Jason Catlin, Brian Schimpf, Pete Moran, Noah Zych,
  et~al.
\newblock Team cornell's skynet: Robust perception and planning in an urban
  environment.
\newblock \emph{Journal of Field Robotics}, 25\penalty0 (8):\penalty0 493--527,
  2008.

\bibitem[Montemerlo et~al.(2008)Montemerlo, Becker, Bhat, Dahlkamp, Dolgov,
  Ettinger, Haehnel, Hilden, Hoffmann, Huhnke, et~al.]{montemerlo2008junior}
Michael Montemerlo, Jan Becker, Suhrid Bhat, Hendrik Dahlkamp, Dmitri Dolgov,
  Scott Ettinger, Dirk Haehnel, Tim Hilden, Gabe Hoffmann, Burkhard Huhnke,
  et~al.
\newblock Junior: The stanford entry in the urban challenge.
\newblock \emph{Journal of field Robotics}, 25\penalty0 (9):\penalty0 569--597,
  2008.

\bibitem[Moreau et~al.(2019{\natexlab{a}})Moreau, Melchior, Victor, Cassany,
  Moze, Aioun, and Guillemard]{moreau2019reactive}
Julien Moreau, Pierre Melchior, St{\'e}phane Victor, Louis Cassany, Mathieu
  Moze, Fran{\c{c}}ois Aioun, and Franck Guillemard.
\newblock Reactive path planning in intersection for autonomous vehicle.
\newblock \emph{IFAC-PapersOnLine}, 52\penalty0 (5):\penalty0 109--114,
  2019{\natexlab{a}}.

\bibitem[Moreau et~al.(2019{\natexlab{b}})Moreau, Melchior, Victor, Moze,
  Aioun, and Guillemard]{moreau2019reactive-be}
Julien Moreau, Pierre Melchior, St{\'e}phane Victor, Mathieu Moze,
  Fran{\c{c}}ois Aioun, and Franck Guillemard.
\newblock Reactive path planning for autonomous vehicle using b{\'e}zier curve
  optimization.
\newblock In \emph{2019 IEEE Intelligent Vehicles Symposium (IV)}, pages
  1048--1053. IEEE, 2019{\natexlab{b}}.

\bibitem[Murray and Sastry(1993)]{murray1993nonholonomic}
Richard~M Murray and Sosale~Shankara Sastry.
\newblock Nonholonomic motion planning: Steering using sinusoids.
\newblock \emph{IEEE transactions on Automatic Control}, 38\penalty0
  (5):\penalty0 700--716, 1993.

\bibitem[{Nilsson} et~al.(2016){Nilsson}, {Hussien}, {Balkan}, {Chen}, {Ames},
  {Grizzle}, {Ozay}, {Peng}, and {Tabuada}]{ACCTwoApproaches}
P.~{Nilsson}, O.~{Hussien}, A.~{Balkan}, Y.~{Chen}, A.~D. {Ames}, J.~W.
  {Grizzle}, N.~{Ozay}, H.~{Peng}, and P.~{Tabuada}.
\newblock Correct-by-construction adaptive cruise control: Two approaches.
\newblock \emph{IEEE Transactions on Control Systems Technology}, 2016.

\bibitem[Paden et~al.(2016)Paden, {\v{C}}{\'a}p, Yong, Yershov, and
  Frazzoli]{paden2016survey}
Brian Paden, Michal {\v{C}}{\'a}p, Sze~Zheng Yong, Dmitry Yershov, and Emilio
  Frazzoli.
\newblock A survey of motion planning and control techniques for self-driving
  urban vehicles.
\newblock \emph{IEEE Transactions on intelligent vehicles}, 1\penalty0
  (1):\penalty0 33--55, 2016.

\bibitem[Park et~al.(2014)Park, Lee, and Han]{park2014development}
Myung-Wook Park, Sang-Woo Lee, and Woo-Yong Han.
\newblock Development of lateral control system for autonomous vehicle based on
  adaptive pure pursuit algorithm.
\newblock In \emph{2014 14th International Conference on Control, Automation
  and Systems (ICCAS)}, pages 1443--1447. IEEE, 2014.

\bibitem[Pendleton et~al.(2017)Pendleton, Andersen, Du, Shen, Meghjani, Eng,
  Rus, and Ang]{machines5010006}
Scott~Drew Pendleton, Hans Andersen, Xinxin Du, Xiaotong Shen, Malika Meghjani,
  You~Hong Eng, Daniela Rus, and Marcelo~H. Ang.
\newblock Perception, planning, control, and coordination for autonomous
  vehicles.
\newblock \emph{Machines}, 5\penalty0 (1), 2017.
\newblock ISSN 2075-1702.
\newblock \doi{10.3390/machines5010006}.

\bibitem[Perez et~al.(2012)Perez, Platt, Konidaris, Kaelbling, and
  Lozano-Perez]{perez2012lqr}
Alejandro Perez, Robert Platt, George Konidaris, Leslie Kaelbling, and Tomas
  Lozano-Perez.
\newblock Lqr-rrt*: Optimal sampling-based motion planning with automatically
  derived extension heuristics.
\newblock In \emph{2012 IEEE International Conference on Robotics and
  Automation}, pages 2537--2542. IEEE, 2012.

\bibitem[P{\^e}tr{\`e}s et~al.(2011)P{\^e}tr{\`e}s, Romero-Ramirez, and
  Plumet]{petres2011reactive}
Cl{\'e}ment P{\^e}tr{\`e}s, Miguel-Angel Romero-Ramirez, and Fr{\'e}d{\'e}ric
  Plumet.
\newblock Reactive path planning for autonomous sailboat.
\newblock In \emph{2011 15th International Conference on Advanced Robotics
  (ICAR)}, pages 112--117. IEEE, 2011.

\bibitem[Prabhakar et~al.(2015)Prabhakar, Duggirala, Mitra, and
  Viswanathan]{prabhakar2015hybrid}
Pavithra Prabhakar, Parasara~Sridhar Duggirala, Sayan Mitra, and Mahesh
  Viswanathan.
\newblock Hybrid automata-based cegar for rectangular hybrid systems.
\newblock \emph{Formal Methods in System Design}, 46\penalty0 (2):\penalty0
  105--134, 2015.

\bibitem[Raffo et~al.(2009)Raffo, Gomes, Normey-Rico, Kelber, and
  Becker]{raffo2009predictive}
Guilherme~V Raffo, Guilherme~K Gomes, Julio~E Normey-Rico, Christian~R Kelber,
  and Leandro~B Becker.
\newblock A predictive controller for autonomous vehicle path tracking.
\newblock \emph{IEEE transactions on intelligent transportation systems},
  10\penalty0 (1):\penalty0 92--102, 2009.

\bibitem[Redding et~al.(2007)Redding, Amin, Boskovic, Kang, Hedrick, Howlett,
  and Poll]{redding2007real}
Joshua Redding, Jayesh Amin, Jovan Boskovic, Yeonsik Kang, Karl Hedrick, Jason
  Howlett, and Scott Poll.
\newblock A real-time obstacle detection and reactive path planning system for
  autonomous small-scale helicopters.
\newblock In \emph{AIAA Guidance, Navigation and Control Conference and
  Exhibit}, page 6413, 2007.

\bibitem[Sharma et~al.(2012)Sharma, Saunders, and Beard]{sharma2012reactive}
Rajnikant Sharma, Jeffery~B Saunders, and Randal~W Beard.
\newblock Reactive path planning for micro air vehicles using bearing-only
  measurements.
\newblock \emph{Journal of Intelligent \&amp; Robotic Systems}, 65\penalty0
  (1):\penalty0 409--416, 2012.

\bibitem[Shkolnik et~al.(2009)Shkolnik, Walter, and
  Tedrake]{shkolnik2009reachability}
Alexander Shkolnik, Matthew Walter, and Russ Tedrake.
\newblock Reachability-guided sampling for planning under differential
  constraints.
\newblock In \emph{2009 IEEE International Conference on Robotics and
  Automation}, pages 2859--2865. IEEE, 2009.

\bibitem[Snider et~al.(2009)]{Snider.2009}
Jarrod~M Snider et~al.
\newblock Automatic steering methods for autonomous automobile path tracking.
\newblock \emph{Robotics Institute, Pittsburgh, PA, Tech. Rep. CMU-RITR-09-08},
  2009.

\bibitem[Takahashi and Schilling(1989)]{takahashi1989motion}
Osamu Takahashi and Robert~J Schilling.
\newblock Motion planning in a plane using generalized voronoi diagrams.
\newblock \emph{IEEE Transactions on robotics and automation}, 5\penalty0
  (2):\penalty0 143--150, 1989.

\bibitem[Talvala et~al.(2011)Talvala, Kritayakirana, and
  Gerdes]{talvala2011pushing}
Kirstin~LR Talvala, Krisada Kritayakirana, and J~Christian Gerdes.
\newblock Pushing the limits: From lanekeeping to autonomous racing.
\newblock \emph{Annual Reviews in Control}, 35\penalty0 (1):\penalty0 137--148,
  2011.

\bibitem[Thrun et~al.(2006)Thrun, Montemerlo, Dahlkamp, Stavens, Aron, Diebel,
  Fong, Gale, Halpenny, Hoffmann, et~al.]{thrun2006stanley}
Sebastian Thrun, Mike Montemerlo, Hendrik Dahlkamp, David Stavens, Andrei Aron,
  James Diebel, Philip Fong, John Gale, Morgan Halpenny, Gabriel Hoffmann,
  et~al.
\newblock Stanley: The robot that won the darpa grand challenge.
\newblock \emph{Journal of field Robotics}, 23\penalty0 (9):\penalty0 661--692,
  2006.

\bibitem[Tiwari and Khanna(2002)]{tiwari2002series}
Ashish Tiwari and Gaurav Khanna.
\newblock Series of abstractions for hybrid automata.
\newblock In \emph{International Workshop on Hybrid Systems: Computation and
  Control}, pages 465--478. Springer, 2002.

\bibitem[Urmson et~al.(2008)Urmson, Anhalt, Bagnell, Baker, Bittner, Clark,
  Dolan, Duggins, Galatali, Geyer, et~al.]{urmson2008autonomous}
Chris Urmson, Joshua Anhalt, Drew Bagnell, Christopher Baker, Robert Bittner,
  MN~Clark, John Dolan, Dave Duggins, Tugrul Galatali, Chris Geyer, et~al.
\newblock Autonomous driving in urban environments: Boss and the urban
  challenge.
\newblock \emph{Journal of Field Robotics}, 25\penalty0 (8):\penalty0 425--466,
  2008.

\bibitem[Vasile and Belta(2014)]{vasile2014reactive}
Cristian~Ioan Vasile and Calin Belta.
\newblock Reactive sampling-based temporal logic path planning.
\newblock In \emph{2014 IEEE International Conference on Robotics and
  Automation (ICRA)}, pages 4310--4315. IEEE, 2014.

\bibitem[Vasile et~al.(2020)Vasile, Li, and Belta]{vasile2020reactive}
Cristian~Ioan Vasile, Xiao Li, and Calin Belta.
\newblock Reactive sampling-based path planning with temporal logic
  specifications.
\newblock \emph{The International Journal of Robotics Research}, 39\penalty0
  (8):\penalty0 1002--1028, 2020.

\bibitem[Xu et~al.(2012)Xu, Wei, Dolan, Zhao, and Zha]{xu2012real}
Wenda Xu, Junqing Wei, John~M Dolan, Huijing Zhao, and Hongbin Zha.
\newblock A real-time motion planner with trajectory optimization for
  autonomous vehicles.
\newblock In \emph{2012 IEEE International Conference on Robotics and
  Automation}, pages 2061--2067. IEEE, 2012.

\end{thebibliography}
